\pdfoutput=1

\documentclass[11pt]{article}

\usepackage[]{acl}

\usepackage{times}
\usepackage{latexsym}

\usepackage[T1]{fontenc}

\usepackage[utf8]{inputenc}

\usepackage{microtype}

\usepackage{inconsolata}

\usepackage{amsthm,amsmath,amsfonts,bm,xspace}
\usepackage{color}
\usepackage[english]{babel}

\newtheorem{corollary}{Corollary}

\def\eg{{\em e.g.,}\xspace}
\def\ie{{\em i.e.,}\xspace}
\def\vs{{\em v.s.}\xspace}
\newcommand{\comment}[1]{}

\usepackage{amsthm,amsmath,amsfonts,bm,xspace}
\usepackage{color}

\def\eg{{\em e.g.,}\xspace}
\def\ie{{\em i.e.,}\xspace}
\def\vs{{\em v.s.}\xspace}

\definecolor{mgreen}{rgb}{0,0.7,0}

\def\eqref#1{(\ref{#1})}

\def\1{\bm{1}}

\def\va{{\bm{a}}}

\def\vc{{\bm{c}}}

\def\ve{{\bm{e}}}

\def\vh{{\bm{h}}}

\def\vm{{\bm{m}}}

\def\vp{{\bm{p}}}
\def\vq{{\bm{q}}}

\def\vs{{\bm{s}}}

\def\vv{{\bm{v}}}

\def\vy{{\bm{y}}}

\def\mH{{\bm{H}}}

\DeclareMathAlphabet{\mathsfit}{\encodingdefault}{\sfdefault}{m}{sl}
\SetMathAlphabet{\mathsfit}{bold}{\encodingdefault}{\sfdefault}{bx}{n}

\def\gS{{\mathcal{S}}}

\def\gX{{\mathcal{X}}}
\def\gY{{\mathcal{Y}}}

\newcommand{\softmax}{\mathrm{softmax}}

\newcommand{\dep}{\not\!\perp\!\!\!\perp}

\newcommand{\system}[1]{\textsc{#1}}

\newcommand{\alpaca}{\system{Alpaca-7B}\xspace}
\newcommand{\chatgpt}{\system{ChatGPT}\xspace}
\newcommand{\chatglm}{\system{ChatGLM}\xspace}
\newcommand{\bart}{\system{BART}\xspace}
\newcommand{\bert}{\system{BERT}\xspace}
\newcommand{\chatyuan}{\system{ChatYuan}\xspace}
\newcommand{\eda}{\system{EDA}\xspace}

\newcommand{\uda}{\system{UDA}\xspace}
\newcommand{\pgb}{\system{PGB}\xspace}
\newcommand{\ourmethod}{\system{IMO}\xspace}

\usepackage{hyperref}
\usepackage{url}
\usepackage{xspace,mfirstuc,tabulary}
\usepackage{microtype}
\usepackage{graphicx}
\usepackage{wrapfig}
\usepackage{amsmath}
\usepackage{subcaption}
\usepackage{enumitem}
\usepackage{svg}

\title{\ourmethod: Greedy Layer-Wise Sparse Representation Learning for Out-of-Distribution Text Classification with Pre-trained Models}

\author{Tao Feng , Lizhen Qu , Zhuang Li , Haolan Zhan , Yuncheng Hua , Gholamreza Haffari \\
        Monash University, Australia \\
        \texttt{\{firstname.lastname\}@monash.edu}}

\begin{document}
\maketitle
\begin{abstract}
Machine learning models have made incredible progress, but they still struggle when applied to examples from unseen domains. This study focuses on a specific problem of domain generalization, where a model is trained on one source domain and tested on multiple target domains that are unseen during training.  
 We propose \ourmethod: \textbf{I}nvariant features \textbf{M}asks for \textbf{O}ut-of-Distribution text classification, to achieve OOD generalization by learning invariant features. During training, \ourmethod would learn sparse mask layers to remove irrelevant features for prediction, where the remaining features keep invariant. Additionally, \ourmethod has an attention module at the token level to focus on tokens that are useful for prediction. Our comprehensive experiments show that \ourmethod substantially outperforms strong baselines such as prompt-based methods and large language models, in terms of various evaluation metrics and settings. 
\end{abstract}

\section{Introduction}
\label{sec:intro}
When deploying natural language processing (NLP) models trained on labeled data in the wild, it is well known that they suffer from poor predictive performance on the samples drawn from the distributions different than their training ~\citep{wang2022surveyDomainGeneralizing}. 
Although various domain adaptation (DA) methods have been proposed~\citep{liu2022deepUDAReview,saunders2022DASurveyMT},  
%
they assume the availability of labeled/unlabeled data from target domains and/or the target domain information.  
%
%
However, for many real-world applications, especially for early-stage businesses, users may apply their models to arbitrary data so the test data may well be Out-of-Distribution (OOD). Hence the domain information may not be available for DA. In addition,  training datasets are often expensive to acquire so  they are available only in one domain. Therefore, this work focuses on single-source \textit{domain generalization} (DG) for text classification, which aims to enable classifiers trained in \textit{one} source domain to \textit{robustly} work on the same classification tasks in any unseen OOD data without any model tuning.

Pre-trained large language models (LLMs) have drawn a lot of attentions due to their strong predictive performance on a variety of tasks. Although generative models or classifiers built on top of pre-trained LLMs outperform prior models in multiple domains, their performance is still not \textit{robust} on tasks when the testing distribution differs substantially from the training distribution~\citep{bang2023multitask}. Recent works~\citep{wang2020CounteringSpuriousCorrelations, 10.1162/tacl_a_00561,veitch2021counterfactual} show that one of the key reasons is \textit{spurious correlations}, which refer to the correlations between features and model outputs that are not based on causal relationships.

To take a step towards ``train it once, apply it anywhere'', we propose a novel greedy layer-wise \textbf{I}nvariant \textbf{M}asking technique for \textbf{O}OD text classification, coined \ourmethod, which selects domain-invariant features and key token representations from appropriate layers of a pre-trained deep transformer encoder to mitigate spurious correlations. The resulting hidden representations are sparse from the top layer to a specific layer of the pretrained model. 
We demonstrate the effectiveness of this technique through theoretical justifications and extensive experiments. Similar to \citep{zhang2022SubnetworkOOD} on computer vision tasks, we shed light on how to apply sparsity as an effective inductive bias to deep pre-trained models for OOD text classification. Our contributions are: 

\begin{itemize}
    \item We propose \ourmethod, a novel top-down greedy layer-wise sparse representation learning method for pre-trained text encoders for robust OOD  classification by sharply reducing task-specific spurious correlations. In comparison with bottom-up layer-wise and simultaneous search across all layers, we discover that the top-down greedy search is decisive for performance improvement.
    \item We develop a theoretical framework that elucidates the relationship between domain-invariant features and causal features. Additionally, we provide an explanation of how our method learns invariant features. 
     \item Our comprehensive experimental results show that:  
        (i) using \ourmethod with \bart~\cite{lewis-etal-2020-bart} significantly outperforms competitive baselines, including \chatgpt, on the classification of topics and sentiment polarity in the majority of the target domains, while \chatgpt has 10 times more parameters than \bart;
        (ii) using  \ourmethod with \chatyuan~\cite{chatyuan} for Chinese  achieves better performance over strong competitors, \eg \chatgpt, on social factor classification;
        (iii) \ourmethod achieves similar OOD performance w.r.t. varying size of training data. The differences of accuracy between using 1k and 3.5 million training instances for models trained with IMO are less than 6\%, and for those trained without IMO is more than 16\%. \footnote{Codes are available at \url{https://github.com/WilliamsToTo/IMO}.}
        

\end{itemize}

\section{Related Work}
\textbf{Domain Generalization.}
Numerous DG methods have been proposed in the past decade, and most of them are designed for multi-source DG~\cite{10.1007/978-3-030-58545-7_18, 10.5555/3495724.3496934, ding2022domain, zhang2022towards, lv2022causality}. Existing DG methods can be roughly classified into two categories: invariant representation learning and data augmentation. The key idea of the former is to reduce the discrepancy between representations of  source domains~\cite{10.5555/3042817.3042820, 8578664, 10.1007/978-3-030-01267-0_38, 8953226, arjovsky2020invariant}. The key idea of data augmentation is to generate out-of-distribution samples, which are used to train the neural network with original source samples to improve the generalization ability~\cite{10.5555/3495724.3496249, wei-zou-2019-eda, volpi2019addressing}. 

This paper focuses on single-source DG, where the model is trained on a single source domain, then evaluated on multiple unseen domains. \citet{wang2021learning} proposes a style-complement module to synthesize images with unseen styles, which are out of original distributions. \citet{qiao2020learning} proposes adversarial domain augmentation to encourage semantic consistency between the augmented and source images in the latent space. \citet{9961940} uses a causality-inspired data augmentation approach to encourage network learning domain-invariant features. In terms of text classification, \citet{10.1162/tacl_a_00468,jia-zhang-2022-prompt}  apply prompt-based learning methods to generate a prompt for each sample, then use large language models to predict labels. 

\noindent
\textbf{Causal Representation Learning (CRL).}
CRL addresses OOD generalization by exploring causal features that lead to labels. It is based on the assumption that causal features are stable across different environments or data selections. Since CRL is very ambitious and even infeasible in real application, a more practical method is invariant representation learning. \citet{10.2307/44682904} investigated that invariant features, to some extent, infer the causal structure. \citet{arjovsky2020invariant} also assumes that prediction conditioned on invariant features is stable under different environments. Following such assumption, a strand of methods tries to learn invariant features by mitigating spurious correlated features, which vary across environments \cite{10.5555/3042817.3042820, 10.1007/978-3-030-58545-7_18, asgari2022masktune, NEURIPS2022_fb64a552, NEURIPS2022_4a9eaf6d}.  This paper also follows this thread of methods, where we treat features that don't affect prediction as spurious correlated features. 


\section{Learning Sparse Domain-Invariant Representations}
LLMs are pre-trained on large-scale corpora so that they can capture rich correlations between tokens across various domains. To enable trained models incorporating LLMs to work across domains, our key idea originates from the \textit{Invariance Assumption} that the conditional distributions of labels conditioned on invariant features do not change across domains~\citep{10.2307/44682904}. \citet{zhang2022SubnetworkOOD} show that the assumption can hold, and there is a subnetwork inside a full network that can achieve better OOD performance than the full network. For a specific classification task, such as sentiment polarity analysis, the assumption indicates that there are certain sparse representations that are potential \textit{causes} of labels~\citep{wang2022representation} across domains. Our method \ourmethod realizes this idea by constructing sparse domain-invariant representations from the hidden representations of the selected layers of pre-trained transformer-based encoders. 

Let $\gX$ be the input space and $\gY$ be the label space, a \textit{domain} is characterized by a joint distribution $P_{XY}$ on $\gX \times \gY$. In the context of a single source DG, we have access to the data of one source domain $\gS = \{(x^{s}, y^{s})\}$ drawn from its joint distribution $P^{\gS}_{XY}$.  The goal is to learn a predictive model $f: \gX \rightarrow \gY$ using only the data sampled from $P^{\gS}_{XY}$ to minimize the prediction error on $K$ unseen target domains, 
each of which is associated with a joint distribution $P^{k}_{XY}$. Due to domain drifts, $P^{\gS}_{XY} \neq P^{k}_{XY}, \forall k \in {1,...,K}$. 

Following~\citep{quinzan2023drcfs}, 
we make the same assumptions that (i) $Y = f(\text{Pa}(Y)) + \epsilon$, where $\text{Pa}(Y)$ denote the features that directly cause $Y$, (ii) $\epsilon$ is exogenous noise, independent of any features, and (iii) $Y$ has no direct causal effect on any features because classification labels are assigned after observing the corresponding texts. Although $P^{\gS}_{XY} \neq P^{(k)}_{XY}, \forall k \in {1,...,K}$, we show in \S \ref{sec:theory} that under all above assumptions, there is a sparse representation $\mathbf{H}_i$ such that the function $Y = f(\mathbf{H}_i) + \epsilon$ exists in both source and target domains. We  empirically study the presence of invariant representations and influence of spurious correlations in \S \ref{sec:analysis}.

\begin{figure}[h]
    \centering
    \includegraphics[width=0.8\columnwidth]{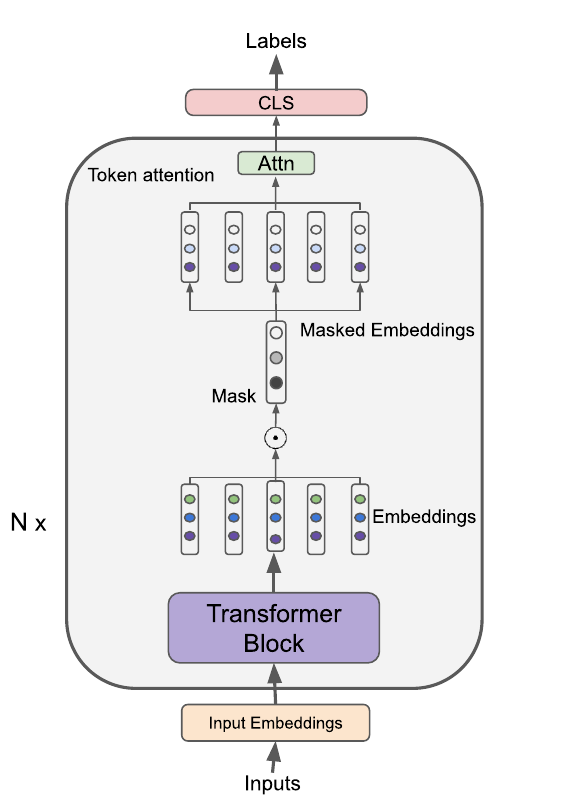}
    \caption{The overall architecture of our method \ourmethod. }
    \label{fig:disentangle_cls_architecture_figure}
    \vspace{-15pt}
\end{figure}

As illustrated in Figure~\ref{fig:disentangle_cls_architecture_figure}, our method constructs sparse domain-invariant representations at both feature and token levels in a top-down manner. At the feature level, given embeddings produced by the transformer block of the top layer, a parametric mask layer identifies invariant features from the embeddings. Then, the mask layer is frozen and the algorithm learns the mask layer for the lower layer. The process is repeated until a pre-specified layer is reached. 
At the token level, a soft attention mechanism incorporates the selected features from the top layer to identify the tokens strongly correlated with $Y$ and use attention weights to create aggregated sparse representations based on the selected features for binary classification. For multi-class classification tasks, a sparse representation is created for each class so that each of them can focus on class-specific information. The model is regularized during training to increase the divergences of the representations between classes.  

\subsection{Extraction of Invariant Features}
Given a text input $X=[x_i]_{i=0}^{T}$, where $x_i$ is a token in $X$, a transformer-based pre-trained language model is employed to convert $x_i$ to a continuous token representation. We use hidden states produced by each transformer layer $l$ as token representations, denoted as $\mH^{l} = [\vh^{l}_{i}]_{i=0}^{T}$. $\vh^{l}_{i}$ embeds both invariant features (useful for prediction in different domains) and spuriously correlated features (irrelevant for prediction) produced by layer $l$. Based on the Invariance Assumption, the invariant features $\vh^{*}$ ensure 
$p^k(Y | \vh^{*})$ to be the same for each domain $k$. 
In a transformer layer $l$, the spuriously correlated features are filtered out by performing element-wise multiplication between token representation $\vh^{l}_{i}$ and a learnable mask $\vm^{l}$. 


%
A parametric filtering vector $\mathbf{m} = \mathbf{r} \odot \mathbf{q}$ contains zero and non-zero elements, where we define a trainable weight vector $\mathbf{r} \in \mathbb{R}^{d}$ and a trainable pruning threshold vector $\mathbf{s} \in \mathbb{R}^{d}$. A unit step function
$
g(t) = 
\begin{cases} 
0 & \text{if } t < 0 \\
1 & \text{if } t \geq 0 
\end{cases}
$
is applied to get a binary mask $\mathbf{q} = g(\left| \mathbf{r} \right| - \mathbf{s})$. By applying element-wise multiplication $\mathbf{e}^{l}_{i} = \mathbf{h}^{l}_{i} \odot \mathbf{m}^{l}$, the zero elements of $\mathbf{m}$ remove corresponding features in token embeddings $\mathbf{h}^{l}$, while non-zero elements characterize the importance of corresponding features. 

As the unit step function $g$ is not differentiable, we approximate its derivative by using the derivative estimator proposed in \cite{DBLP:conf/bmvc/XuC19} such that all parameters of a mask layer are trainable by using back-propagation and the family of stochastic gradient descent algorithms,

\vspace{-2mm}
\begin{small}
    \begin{equation}
\label{equ: long_tailed estimator}
    \frac{d}{dt}g(t) = 
    \begin{cases}
        2 - 4|t|, & -0.4\leq t \leq 0.4 \\ 
        0.4, & 0.4 \leq |t| \leq 1\\ 
        0, & \text{otherwise.}
    \end{cases}
\end{equation}
\end{small}
\noindent Following~\cite{DBLP:conf/bmvc/XuC19}, we add a sparse regularization term $L_{sparse}$ to the training loss to encourage the sparsity of mask layers:

\vspace{-2mm}
\begin{small}
    \begin{equation}
\label{equ: sparse_regularizer}
\mathcal{L}_{sparse} = \sum_{i=1}^{N} \exp(-\vs_{i}), \vs \in \mathbb{R}^{d} 
\end{equation}
\end{small}
\noindent where $\exp(-\vs_{i})$ encourages high  (but not extremely large) thresholds. A higher threshold leads to removal of more features. 
During \textit{inference}, we retain the mask layers to retain invariant features while discarding irrelevant ones.

\subsection{Identification of Invariant Tokens}
Given a long token sequence, not all information is useful for target tasks. For example, function words, such as `the', or `that', provide little information for predicting sentiment polarity. Thus, we employ a token-level attention mechanism to focus on important tokens. Instead of using all features of a token representation, we compute attention scores by using only the  invariant features. The proposed attention mechanism differs slightly between binary and multi-class classification.

\paragraph{Binary Classification.}
For binary classification, we treat the filtering vector $\mathbf{m}^{L}$ from the last layer $L$ as the query vector and compute the attention weight by performing the matrix product between $\mathbf{m}^{L}$ and each token embedding from the last layer $\mathbf{e}_i^{L}$: $a_i = \mathbf{m}^{L} \mathbf{e}_i^{L}$.
Here, the filtering vector and token embeddings are interpreted as matrices, with $\mathbf{m}^{L} \in \mathbb{R}^{1 \times d}$ and $\mathbf{e}_i^{L} \in \mathbb{R}^{d \times 1}$. For an input token sequence, we aggregate the masked token embeddings to obtain a sequence representation $\mathbf{v} = \sum_{i}^{T} a_{i} \mathbf{e}_{i}^{L}$, where $\mathbf{v}\in \mathbb{R}^{1 \times d}$. Finally, the sequence representation is fed into a fully-connected layer, followed by generating a distribution over the label space as follows: $\hat{\mathbf{y}} = \softmax(\mathbf{v} \mathbf{P})$.

\paragraph{Multi-class Classification.}
For the multi-class classification task, we propose using multiple mask layers $\vm^{L}_{y}$ in the last layer $L$ to capture corresponding features and tokens for  labels $\vy$. The number of mask layers equals the number of labels. Each label has its own attention weights $\va^{L}_{y} = \vm^{L}_{y} \ve$, and its own representation $\vv^{L}_{y} = \sum_{i}^{T} a^{L}_{yi}  \ve_{i}$. Instead of using a fully-connected layer, we use a learnable weight vector per class to project $\vv^{L}$ to a scalar: $c^{L} = \vv^{L} \vp^{L}$, where $\vv^{L} \in \mathbb{R}^{1\times d}$ and $\vp^{L} \in \mathbb{R}^{d\times 1}$. The rationale behind this is that each class should have its own weight vector and hidden representations for encoding class-specific information. Then, we concatenate these scalars to a vector $\vc = [c^{L}]$, and compute the predictive distribution by $\hat{\vy} = \softmax (\vc)$.

To encourage mask layers to extract label-specific features, we propose the following regularization term to penalize pairwise cosine similarities between the corresponding mask layers (where $N$ is the number of label-specific mask layers):
\begin{equation}
    \mathcal{L}_{dist} = \frac{1}{N(N-1)}\sum_{i \neq j}\cos(\vm^{i}, \vm^{j}).
\end{equation}

\paragraph{Training Procedure.}
Rather than training all mask layers simultaneously, we adopt a layer-wise training procedure to train them sequentially from the top layer to the bottom layer. As illustrated in Figure~\ref{fig:disentangle_cls_architecture_figure}, for each layer, a new filtering layer, $\mathbf{m}^{L-i}$, is introduced on the top of the $(L-i)$-th transformer layer, with $i \in \{0, 1, 2, ... L-1\}$. Crucially, during this phase, the previously trained mask layers remain frozen to preserve their learned parameters. Upon each layer's training completion, the model is stored as $\theta_{L:L-i}$. This iterative procedure continues until the training of the most bottom filtering vector, $\mathbf{m}^{1}$, is completed. Consequently, a suite of models, ranging from $\theta_{L}$ to $\theta_{L:1}$, is collected. We empirically determine the model's efficacy by evaluating its performance on the validation set from the source domain. The best-performing model is chosen as the model to test on the target domains.

\paragraph{Objective Function.}
During training, the overall objective for binary classification is to (1) have good predictive performance on classification tasks and (2) maximize sparsity in mask layers to only keep invariant features. When training mask at layer $l$, the loss function is:
\begin{equation}
\label{eq:loss}
    \mathcal{L} = \mathcal{L}_{ce} + \alpha \mathcal{L}_{sparsity}^{l}
\end{equation}
where $\mathcal{L}_{ce}$ denotes  the cross entropy loss and $f$ denotes the predictive model. $\alpha$, where $\alpha>0$, is a hyperparameter that controls the balance between predictive performance and sparsity in mask layers. $\mathcal{L}_{sparsity}^{l}$ is the sparse regularization term for mask at layer $l$. 

For multi-class classification, we add a distance regularization term:
\begin{equation}
    \mathcal{L} = \mathcal{L}_{ce} + \alpha \mathcal{L}_{sparsity}^{l} + \beta \mathcal{L}_{dist}
\end{equation}
The hyperparameter $\beta$ serves to calibrate the equilibrium between features specific to individual labels and those shared across multiple labels.


\begin{figure}[t]
\centering
   \begin{subfigure}{0.3\columnwidth}
       \includegraphics[width=\columnwidth]{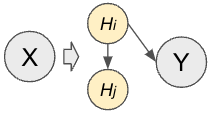}
       \caption{}
       \label{fig:causal_feature_a}
   \end{subfigure}
   \hfill
   \begin{subfigure}{0.3\columnwidth}
       \includegraphics[width=\columnwidth]{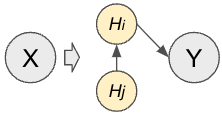}
       \caption{}
       \label{fig:causal_feature_b}
   \end{subfigure}
   \hfill
   \begin{subfigure}{0.3\columnwidth}
       \includegraphics[width=\columnwidth]{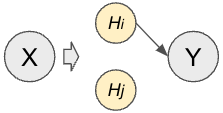}
       \caption{}
       \label{fig:causal_feature_c}
   \end{subfigure}


    
   \caption{Illustration of potential causal graphs between the variables $H_i$, $H_j$ of two features (encoded from an input $X$) and a target variable $Y$. }
   \label{fig:invariant_direct_cause_relation}
   \vspace{-15pt}
\end{figure}

\subsection{Theoretical Analysis}
\label{sec:theory}
Based on our assumptions, $Y = f(\mathbf{H}_i) + \epsilon$ exists, when $\mathbf{H}_i$ are the parent nodes of $Y$ in the underlying causal graph. Because $\mathbf{H}_i$ are a subset among all possible hidden representations correlated with $Y$, there should be a subset of hidden representations serving as parents of $Y$, otherwise the invariance assumption does not hold.  Due to the widely used faithfulness assumption stating that statistical independences imply the corresponding causal structures~\citep{neal2020introduction}, we aim to find out $\mathbf{H}_i \dep Y | \mathbf{H}_j$, where $\mathbf{H}_j$ is any feature set non-overlapped with $\mathbf{H}_i$.  

We start our theoretical analysis by introducing a sparsity regularization term $\Omega(Y, H_i, ..., H_j)$, which counts the number of edges between $Y$ and the random variables of features in an underlying causal graph, where $Y$ is the variable for labels and $H_k$ denotes the random variable of the feature $h_k$. Then we introduce a loss function $\mathcal{L}_{\Omega}(Y, H_i, ..., H_j) = \mathcal{L}_{ce} + \alpha \Omega(Y, H_i, ..., H_j)$, analogous to Eq. \eqref{eq:loss}.

Considering the simplest case that there is only a causal feature $h_i$ and a non-causal feature $h_j$, the corresponding random variables are denoted by $H_i$ and $H_j$. From any causal graphs in Fig. \ref{fig:invariant_direct_cause_relation}, we conclude that $p(Y | H_i, H_j) = p(Y | H_i)$ so that the cross entropy term in $\mathcal{L}_{\Omega}$ remains the same when using the term $p(Y | H_i)$, but the loss decreases after removing the non-causal feature from the loss due to the regularization term $\Omega(Y, H_i, H_j)$.  

The two feature case can be easily extended to the case having more than two features. It is trivial that excluding a non-causal feature from the loss $\mathcal{L}_{\Omega}$ leads to the decrease of $\mathcal{L}_{\Omega}$ due to the Markov property of causal graphs~\citep{10.5555/3202377}.
\begin{corollary}
If there is no edge between $Y$ and $H_k$ in a causal graph, then $\mathcal{L}_{\Omega}(Y, H_i, ..., H_j) < \mathcal{L}_{\Omega}(Y, H_i, ..., H_j, H_k)$. 
\end{corollary}

During training, we start with a loss $\mathcal{L}_{\Omega}(Y, H_1, ..., H_N)$ with a complete set of features. If a non-causal feature $H_k$ is removed, $\mathcal{L}_{\Omega}(Y, H_i, ..., H_j)$ decreases according to Corollary 1. In contrast, if a causal feature $H_k$ is removed, the cross entropy term increases because the mutual information $I(Y; H_k | H_i, ..., H_j) > 0$. Namely, $H_k$ adds additional information for predicting $Y$. However, in that case, $\mathcal{L}_{\Omega}(Y, H_i, ..., H_j)$ may still decrease if the increase of $\mathcal{L}_{ce}$ is smaller than the decrease of the regularization term $\alpha \mathcal{L}_{\Omega}(Y, H_i, ..., H_j)$, where $\alpha>0$. The exceptional case can be mitigated if $\alpha$ is sufficiently small. As a result, the loss $\mathcal{L}_{\Omega}$ provides an effective way to guide the search for the features serving as the causes of the labels, although we cannot recover the underlying true causal graphs. Herein, the loss \eqref{eq:loss} is a surrogate of $\mathcal{L}_{\Omega}(Y, H_i, ..., H_j)$ by using a deep neural network. 

\section{Experiments}
We show that our approach significantly outperforms the competitive baselines in almost all settings, followed by empirically verifying that domain-invariant sparse representations indeed exist and spurious features deteriorate model performance in Sec. \ref{sec:analysis}, as well as justifying the effectiveness of top-down greedy search strategy and individual modules in the ablation study.

\subsection{Experimental Setup}
\paragraph{Tasks and Datasets}
We evaluate our method on binary and multi-class classification tasks. Herein, we adopt \textit{accuracy} as the metric for binary sentiment polarity classification and \textit{macro-F1} for multi-class classification tasks. All models are trained with five different random seeds to assess the statistical significance. 

The datasets for binary sentiment analysis include Amazon Review Polarity \cite{zhang2015character}, Yelp Review Polarity \cite{zhang2015character}, IMDB \cite{maas-EtAl:2011:ACL-HLT2011}, TweetEval Sentiment \cite{barbieri-etal-2020-tweeteval} \footnote{We remove all neural instances to turn it into a binary classification task. } and Yahoo! Answers Sentiment \cite{li-etal-2019-domain}. 
For multi-class classification, we consider topic classification task in AG News dataset~\cite{10.1145/1062745.1062778, 10.1145/1060745.1060764, NIPS2015_250cf8b5} and social factor prediction task in SocialDial \cite{zhan2023socialdial}. More details about datasets can be found in Appendix~\ref{apx:experiment_datasets}.

\begin{table*}[h]
\centering
\resizebox{\textwidth}{!}{%
\begin{tabular}{l||ccc|ccc|ccc|ccc|c}
\hline
\textbf{}        & \multicolumn{3}{c|}{\textbf{IMDB$\rightarrow$}}       & \multicolumn{3}{c|}{\textbf{Amazon$\rightarrow$}}    & \multicolumn{3}{c|}{\textbf{Yelp$\rightarrow$}}       & \multicolumn{3}{c|}{\textbf{TweetEval$\rightarrow$}} &                \\
\textbf{Models}  & \textbf{Amazon} & \textbf{Yelp}  & \textbf{TweetEval} & \textbf{IMDB}  & \textbf{Yelp}  & \textbf{TweetEval} & \textbf{IMDB}  & \textbf{Amazon} & \textbf{TweetEval} & \textbf{IMDB}   & \textbf{Yelp}   & \textbf{Amazon}  & \textbf{Avg.}  \\ \hline \hline
\bert            & 89.77*          & 87.12*         & 78.52*             & 88.09*         & 92.18*         & 83.75*             & 86.98*         & 92.10*          & 87.55*             & 82.59*          & 84.87*          & 86.80*           & 86.69*         \\
\bart            & 89.91*          & 88.01*         & 68.47*             & 87.93*         & 91.01*         & 82.98*             & 86.44*         & 91.97*          & 88.21*             & 78.21*          & 89.51*          & 87.01*           & 85.80*         \\
\bert-\eda       & 87.73*          & 87.47*         & 72.10*             & 88.89*         & 92.43*         & 86.40*             & 88.11*         & 92.98*          & 87.92*             & 81.64*          & 85.82*          & 87.77*           & 86.61*         \\
\bert-\uda       & 87.76*          & 87.02*         & 70.23*             & 89.87*         & 93.78*         & 86.37*             & 86.89*         & 92.81*          & 84.91*             & 82.83*          & 85.95*          & 87.29*           & 86.31*         \\
\bert-\pgb       & 88.40*          & 83.61*         & 70.51*             & 89.70*         & 93.66*         & 86.19*             & 86.09*         & 92.72*          & 87.95*             & 81.88*          & 85.13*          & 87.54*           & 86.11*         \\
PADA             & 85.73*          & 89.84*         & 88.40              & 84.47*         & 93.96          & 85.92*             & 87.71*         & 91.42*          & 90.33              & 80.30*          & 84.69*          & 90.61            & 87.78*         \\
PDA              & 89.35*          & 90.59*         & 87.71*             & 88.16*         & 94.20          & 85.61*             & 88.17*         & 93.59           & 89.88*             & 82.05           & 86.37           & 86.41            & 88.51*         \\
\chatgpt         & 91.08           & 92.06          & 81.01              & 90.50          & 92.06          & 81.01              & \textbf{90.50} & 91.08           & 81.01              & \textbf{90.50}  & 92.06           & 91.08            & 88.66          \\
\alpaca          & 90.14           & 92.30          & 88.66              & 83.01          & 92.30          & 88.66              & 83.01          & 90.14           & 88.66              & 83.01           & 92.30           & 90.14            & 88.52          \\
\alpaca-LoRA     & 89.80           & 82.80          & 87.77              & 81.00          & 82.80          & 87.77              & 81.00          & 89.80           & 87.77              & 81.00           & 82.80           & 89.80            & 85.34          \\ \hline
\ourmethod-\bart (Our) & \textbf{93.97}  & \textbf{94.63} & \textbf{89.58}     & \textbf{90.86} & \textbf{95.14} & \textbf{91.08}     & 90.08          & \textbf{94.87}  & \textbf{91.62}     & 85.39           & \textbf{92.84}  & 91.66            & \textbf{91.81} \\
\ourmethod-\bart B2T             & 75.86*          & 75.37*         & 71.90*             & 73.27*         & 73.74*         & 72.58*             & 72.90*         & 73.47*          & 72.06*             & 69.74*          & 73.29*          & 75.81*           & 73.33*         \\
\ourmethod-\bart w/o \textbf{sq} & 74.88*          & 76.41*         & 67.97*             & 70.47*         & 72.33*         & 71.98*             & 71.59*         & 72.30*          & 71.73*             & 71.25*          & 71.62*          & 70.63*           & 71.93*         \\
\ourmethod-\bart last            & 91.71*          & 92.82*         & 89.01              & 89.41          & 93.01*         & 89.85*             & 89.67          & 93.51           & 90.10*             & 84.69*          & 91.22*          & 90.95*           & 90.49*         \\

\hline
\end{tabular}%
}
\caption{Single-source domain generalization on sentiment analysis datasets. ``B2T'': bottom-up layer-wise search. ``\textbf{w/o sq}'':  simultaneous search. ``last'': applying the mask on only the last layer. 
The metric is accuracy.  Asterisk * shows a significant difference compared to IMG-BART using a t-test with a $p\leq 0.05$. }
\label{tab:main_result_sentiment_cls}
\vspace{-15pt}
\end{table*}

\begin{table}
\centering
\small
\resizebox{\columnwidth}{!}{%
\begin{tabular}{lccc}
\hline
\multicolumn{4}{c}{\textbf{AG News}}                                                                                            \\ \hline
\multicolumn{1}{l|}{\textbf{Models}}  & \textbf{Title $\rightarrow$ Desc} & \textbf{Desc $\rightarrow$ Title} & \textbf{Avg-F1} \\ \hline
\multicolumn{1}{l|}{\bert}            & 81.11*                            & 67.95*                            & 74.68*          \\
\multicolumn{1}{l|}{\bart}            & 80.12*                            & 71.22*                            & 75.96*          \\
\multicolumn{1}{l|}{\bert-\eda}       & 80.52*                            & 72.10*                            & 76.58*          \\
\multicolumn{1}{l|}{\bert-\uda}       & 80.41*                            & 71.81*                            & 75.82*          \\
\multicolumn{1}{l|}{\bert-\pgb}       & 78.53*                            & 73.51*                            & 76,02*          \\
\multicolumn{1}{l|}{PADA}             &   82.39*                          &  75.52*                                 &     78.96*            \\
\multicolumn{1}{l|}{PDA}              &     83.61*                      &  75.96*                                 &     79.79*            \\
\multicolumn{1}{l|}{\chatgpt}         & 85.13                             & 79.28                             & 82.21           \\
\multicolumn{1}{l|}{\alpaca}          & 70.61                             & 70.44                             & 71.49          \\
\multicolumn{1}{l|}{\alpaca-LoRA}     & 56.17                             & 49.44                             & 52.81          \\ \hline
\multicolumn{1}{l|}{\ourmethod-\bart (Our)} & \textbf{89.40}                   & \textbf{81.97}                   & \textbf{85.68} \\
\multicolumn{1}{l|}{\ourmethod-\bart B2T}             & 70.31*                            & 64.59*                            & 67.45*          \\
\multicolumn{1}{l|}{\ourmethod-\bart w/o \textbf{sq}} & 62.59*                            & 57.27*                            & 59.93*          \\
\multicolumn{1}{l|}{\ourmethod-\bart last}            & 88.22                             & 80.05*                            & 84.13*          \\
\hline
\end{tabular}%
}
\caption{Results for multi-class classification datasets. `Desc' represents description. The metric is macro F1.}
\label{tab:main_result_multiclass_cls}
\end{table}
\begin{table}[h]
\centering
\small
\resizebox{\columnwidth}{!}{%
\begin{tabular}{lcccc}
\hline
\multicolumn{5}{c}{\textbf{SocialDial}}                                                                                                                                                                                                                                                                                                                      \\ \hline
\multicolumn{1}{l|}{\textbf{Models}}      & \textbf{\begin{tabular}[c]{@{}c@{}}Loc (Synthetic)  \\ $\rightarrow$  Loc (Human)\end{tabular}} & \textbf{\begin{tabular}[c]{@{}c@{}}SD (Synthetic)  \\ $\rightarrow$  SD(Human)\end{tabular}} & \textbf{\begin{tabular}[c]{@{}c@{}}SR (Synthetic)  \\ $\rightarrow$  SR(Human)\end{tabular}} & \textbf{Avg- F1} \\ \hline
\multicolumn{1}{l|}{\bert-zh}             & 18.11*                                                                                          & 35.05*                                                                                       & 32.39*                                                                                       & 28.51*           \\
\multicolumn{1}{l|}{\chatyuan}            & 18.23*                                                                                          & 34.94*                                                                                       & 33.92*                                                                                       & 29.03*           \\
\multicolumn{1}{l|}{\bert-\eda}           & 13.98*                                                                                          & 35.71*                                                                                       & 26.38*                                                                                       & 25.36*           \\
\multicolumn{1}{l|}{\bert-\uda}           & 15.20*                                                                                          & 33.59*                                                                                       & 27.03*                                                                                       & 25.27*           \\
\multicolumn{1}{l|}{\chatgpt}             & 21.44                                                                                           & 38.46                                                                                        & 35.12                                                                                        & 31.67            \\
\multicolumn{1}{l|}{\chatglm-6B}          & 20.57                                                                                           & 20.53                                                                                        & 11.55                                                                                        & 17.55            \\ \hline
\multicolumn{1}{l|}{\ourmethod-CY (Our)} & \textbf{23.22}                                                                                  & \textbf{46.04}                                                                               & \textbf{42.71}                                                                               & \textbf{37.32}   \\
\multicolumn{1}{l|}{\ourmethod-CY B2T}             & 14.31*                                                                                          & 30.29*                                                                                       & 32.45*                                                                                       & 25.68*           \\
\multicolumn{1}{l|}{\ourmethod-CY w/o \textbf{sq}} & 13.37*                                                                                          & 29.81*                                                                                       & 29.05*                                                                                       & 24.07*            \\
\multicolumn{1}{l|}{\ourmethod-CY last}            & 21.47*                                                                                          & 44.73                                                                                        & 39.89*                                                                                       & 35.36*           \\
\hline
\end{tabular}%
}
\caption{Evaluation results on SocialDial dataset. CY represents the pre-trained language model \chatyuan. Loc represents Location; SD represents Social Distance; SR represents Social Relation. The metric is macro F1.}
\label{tab:main_result_multiclass_socialdial_cls}
\vspace{-15pt}
\end{table}

\paragraph{Baseline Models.}
As \cite{gulrajani2021in} showed, simple empirical risk minimization (ERM) outperforms many SOTA domain generalization algorithms. So we finetune \textbf{\bert} \cite{devlin-etal-2019-bert} and encoder of \textbf{\bart} \cite{lewis-etal-2020-bart} using cross-entropy loss as baselines. For Chinese text classification, we use \textbf{\bert-zh} \cite{devlin-etal-2019-bert}, \textbf{\bart-zh} \cite{shao2021cpt} and \textbf{\chatyuan} \cite{chatyuan}.

\noindent
\textit{\underline{Domain Generalization Models.}}
\textbf{PADA} \cite{10.1162/tacl_a_00468} is an example-based autoregressive prompt learning algorithm for domain generalization based on the T5 language model \cite{10.5555/3455716.3455856}. 
\textbf{PDA} \cite{jia-zhang-2022-prompt} is a prompt-based learning algorithm for domain generalization.

\noindent
\textit{\underline{Large Language Models.}}
As \chatgpt shows promising zero-shot ability on various NLP tasks \cite{openai2023gpt4}, we treat \textbf{\chatgpt} (gpt-3.5-turbo) as a baseline. \textbf{\alpaca} \cite{alpaca} is another baseline, which is finetuned from  LLaMA  7B \cite{touvron2023llama} on 52K instruction-following data generated by self-instruct \cite{wang2022selfinstruct}. \textbf{\alpaca-LoRA} is a finetuned \alpaca model using low-rank adaptation \cite{alpaca-lora,hu2022lora}. \textbf{\chatglm-6B} \cite{chatglm_6b} is an open large language model based on General Language Model \cite{du-etal-2022-glm}, optimized for Chinese question-answering and dialogue. All LLMs use few-shot in-context learning. The specific query templates used for the LLMs can be found in Appendix~\ref{apx:training_details}.
\noindent
\textit{\underline{Data Augmentation.}}
\citet{wiles2022a, gokhale-etal-2022-generalized} find data augmentation benefit domain generalization tasks. \textbf{\eda}~\cite{wei-zou-2019-eda} uses four operations (\ie synonym replacement, random insertion, random swap, and random deletion) to augment text data. \textbf{\uda}~\cite{10.5555/3495724.3496249} uses back-translation to generate diverse paraphrases while preserving the semantics of the original sentences. \textbf{\pgb}~\cite{shiri2023paraphrasing} generates syntactically and lexically diversified paraphrases using a fine-tuned BART.

\begin{figure*}[t]
    \centering
    \includegraphics[width=\textwidth]{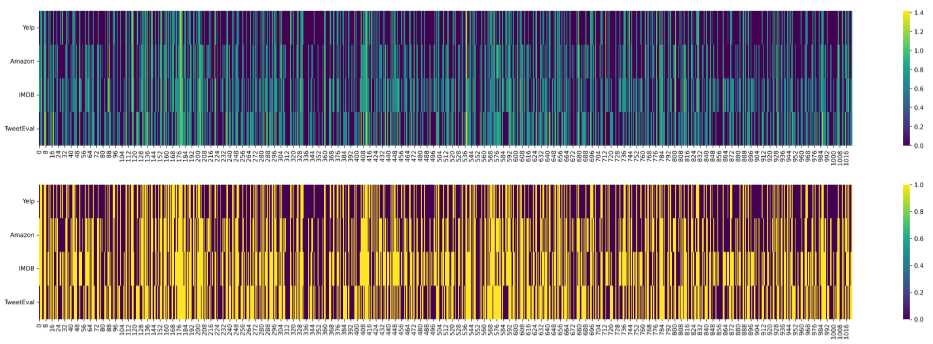}
    \caption{Visualization of filtering and mask vectors in \ourmethod-\bart. The top figure visualizes the filtering vectors $\vm$, while the bottom one visualizes the mask vectors $\vq$. The x-axis signifies the dimensionality of mask layers, whereas the y-axis denotes values attributed to each dimension.}
    \label{fig:mask_layer_visual_bart}
    \vspace{-15pt}
\end{figure*}

\subsection{Domain Generalization Results}

\paragraph{Binary Classification.} Table~\ref{tab:main_result_sentiment_cls} reports the comparisons between our method and the baselines on sentiment polarity classification. Our method using \bart as backbone (\ie \ourmethod-\bart) achieves superior performance over all baselines in 7 of 12 settings, and outperforms the best baseline \chatgpt by 2.63\% on average. Interestingly, \chatgpt stands out as the best model in two out of 12 settings, though it remains unclear whether \chatgpt use those datasets for training. 
Moreover, it is noteworthy that data augmentation methods (\ie \bert-\eda, \bert-\uda, \bert-\pgb) show slightly inferior performance in comparison to the simple fine-tuning of \bert in terms of average accuracy. This suggests that simply back-translating or paraphrasing instances within source domains does not enhance performance on target domains. 

\paragraph{Multi-class Classification.} As shown in Table~\ref{tab:main_result_multiclass_cls} and Table ~\ref{tab:main_result_multiclass_socialdial_cls}, our method outperforms all baselines in terms of average macro-F1 by 3.22\% and 5.16\% on AG News and SocialDial respectively. Among baselines, \chatgpt exhibits the strongest performance on both datasets and surpasses \alpaca, \alpaca-LoRA, and \chatglm by a large margin. This superior performance shows that current open-source large language models still have a substantial performance gap with \chatgpt when handling difficult tasks.

\subsection{Analysis of Spurious Features}
\label{sec:analysis}

\paragraph{Presence of Invariant Representations.} We inspect shared representations at both feature and token levels. Invariant features are expected to have non-zero values across domains. Taking the best performing model \ourmethod-\bart in the sentiment analysis as an example, we train the model in each domain respectively and visualize its masks of the top layer in each domains. As depicted in Fig. \ref{fig:mask_layer_visual_bart}, there are indeed a set of features shared across domains selected by the masks. We further compute Cosine similarities between the filtering vectors $\vm$ of the top layer trained on different source domains. As shown in Table \ref{tab:mask_vector_cosine_similarity} in Appendix, their similarities range from 0.68 and 0.85. At the token level, we inspect the shared attention weights visualized in Fig.\ref{fig:attention_weights_yelp} (see Appendix A3), which indicate the keywords shared across domains in sentiment analysis, such as ``great'' and ``slow''. 

\paragraph{Impact of Spurious Correlations.}
To study whether our proposed masking mechanism indeed identifies robust features, we compare the performance of using the selected features with the non-selected ones. Specifically, we run additional experiments by replacing the learned binary masks $\mathbf{q}$ with $\left| 1-\mathbf{q} \right|$, followed by freezing all parameters except the classification head and training a model using those non-selected features. The results in Table~\ref{tab:study_reverse_mask} show that models using the non-selected features have an approximate 6\% accuracy reduction in source domains and perform worse than using all features. In target domains, the corresponding performance drop using non-selected features is significantly higher than that using both our method as well as using all features. Hence, our masks indeed mitigate the use of spurious features.

\begin{table}[h]
\centering
\resizebox{\columnwidth}{!}{%
\begin{tabular}{l|cccc|cccc}
\hline
\textbf{}                  & \multicolumn{4}{c|}{\textbf{Yelp$\rightarrow$}}                                                                                                                                                                                                                                  & \multicolumn{4}{c}{\textbf{Amazon$\rightarrow$}}                                                                                                                                                                                                                                 \\
\textbf{Models}            & \textbf{\begin{tabular}[c]{@{}c@{}}Yelp\\ (Source)\end{tabular}} & \textbf{\begin{tabular}[c]{@{}c@{}}IMDB\\ (Target)\end{tabular}} & \textbf{\begin{tabular}[c]{@{}c@{}}Amazon\\ (Target)\end{tabular}} & \textbf{\begin{tabular}[c]{@{}c@{}}TweetEval\\ (Target)\end{tabular}} & \textbf{\begin{tabular}[c]{@{}c@{}}Amazon\\ (Source)\end{tabular}} & \textbf{\begin{tabular}[c]{@{}c@{}}IMDB\\ (Target)\end{tabular}} & \textbf{\begin{tabular}[c]{@{}c@{}}Yelp\\ (Target)\end{tabular}} & \textbf{\begin{tabular}[c]{@{}c@{}}TweetEval\\ (Target)\end{tabular}} \\ \hline
\ourmethod           & 95.94                                                            & -5.86                                                            & -1.07                                                              & -4.32                                                                 & 95.34                                                              & -4.48                                                            & -0.20                                                            & -4.26                                                                 \\
\ourmethod - SC & 89.01*                                                           & -7.81*                                                           & -3.88*                                                             & -11.20*                                                               & 90.12*                                                             & -7.64*                                                           & -3.47*                                                           & -12.59*                                                               \\ \hline
\end{tabular}%
}
\caption{Comparison between the proposed model and model using spurious features (SC). In target datasets, we report the reduced percentage of accuracy compared to the source domains.} 
\label{tab:study_reverse_mask}
\vspace{-15pt}
\end{table}

\begin{table*}[h]
\centering
\resizebox{\textwidth}{!}{%
\begin{tabular}{l|ccc|ccc|ccc|ccc|c}
\hline
                                 & \multicolumn{3}{c|}{\textbf{IMDB$\rightarrow$}}       & \multicolumn{3}{c|}{\textbf{Amazon$\rightarrow$}}    & \multicolumn{3}{c|}{\textbf{Yelp$\rightarrow$}}       & \multicolumn{3}{c|}{\textbf{TweetEval$\rightarrow$}} &                \\
\textbf{Models}                  & \textbf{Amazon} & \textbf{Yelp}  & \textbf{TweetEval} & \textbf{IMDB}  & \textbf{Yelp}  & \textbf{TweetEval} & \textbf{IMDB}  & \textbf{Amazon} & \textbf{TweetEval} & \textbf{IMDB}   & \textbf{Yelp}   & \textbf{Amazon}  & \textbf{Avg.}  \\ \hline
\bart w/o \ourmethod    & 89.94*          & 89.13*         & 69.59*             & 88.19*         & 92.20*         & 82.69*             & 86.85*         & 90.64*          & 85.83*             & 78.98*          & 89.25*          & 87.58*           & 85.91*         \\
\ourmethod-\bart (Our)                & \textbf{93.97}  & \textbf{94.63} & \textbf{89.58}     & \textbf{90.86} & \textbf{95.14} & \textbf{91.08}     & \textbf{90.08} & \textbf{94.87}  & \textbf{91.62}     & \textbf{85.39}  & \textbf{92.84}  & \textbf{91.66}   & \textbf{91.81} \\
\ourmethod-\bart w/o $\vm$       & 92.15*          & 92.49*         & 85.61*             & 89.48*         & 92.97*         & 88.53*             & 88.28          & 92.75           & 87.44              & 80.10           & 89.57           & 88.09*           & 88.95*         \\
\ourmethod-\bart w/o $\va$       & 91.35*          & 91.04*         & 84.18*             & 88.51*         & 92.49*         & 84.97*             & 87.10*         & 91.87*          & 88.01*             & 83.31*          & 90.61*          & 88.87*           & 88.52*         \\
\ourmethod-\bart STE             & 91.11*          & 91.71*         & 88.05*             & 88.29*         & 91.69*         & 87.09*             & 88.91*         & 91.39*          & 89.12*             & 82.48*          & 89.37*          & 88.50*           & 88.97*         \\
\ourmethod-\bart STR             & 89.79*          & 88.97*         & 72.98*             & 86.26*         & 87.48*         & 79.48*             & 86.40*         & 88.31*          & 77.49*             & 81.43*          & 85.13*          & 82.49*           & 83.85*         \\
\ourmethod-\bart Scalar          & 87.31*          & 89.92*         & 87.34*             & 87.73*         & 86.03*         & 83.41*             & 87.11*         & 86.43*          & 85.94*             & 81.44*          & 84.75*          & 85.41*           & 86.06*         \\ 
T5 w/o \ourmethod   & 87.53*           & 87.09*          & 66.37*             & 86.47*         & 89.38*         & 80.68*             & 84.94*         & 88.86*          & 83.78*              & 76.48*          & 86.89*           & 85.53*           & 83.67*         \\
\ourmethod-T5    & 93.45           & 93.88          & 84.92*             & 89.23*         & 93.38*         & 89.73*             & 88.27*         & 93.02*          & 91.01              & 81.39*          & 91.93           & 89.97*           & 90.01*         \\
\bert  w/o \ourmethod & 86.48*          & 86.19*         & 66.28*             & 86.12*         & 88.91*         & 81.45*             & 86.34*         & 88.34*          & 83.46*             & 77.25*          & 87.34*          & 84.82*           & 83.58*         \\
\ourmethod-\bert & 92.09*          & 91.93*         & 85.34*             & 88.53*         & 92.19*         & 88.17*             & 87.46*         & 91.49*          & 89.55*             & 79.93*          & 89.23*          & 87.73*           & 88.64*         \\ 

\hline
\end{tabular}%
}
\caption{Ablation study on sentiment analysis datasets. }
\label{tab:ablation_study_sentiment_cls}
\vspace{-15pt}
\end{table*}

\subsection{Ablation Study}
We compare top-down greedy search with alternative methods: bottom-up layer-wise search (B2T), simultaneous search (w/o \textbf{sq}), and only applying a mask on the last layer (last). From Table~\ref{tab:main_result_sentiment_cls},~\ref{tab:main_result_multiclass_cls} and ~\ref{tab:main_result_multiclass_socialdial_cls}, we can tell that top-down greedy performs significantly better than the alternative competitors. We conjecture that top-down layer-wise learning serves a regularization method that reduces the risk of loosing crucial features that are well correlated with $Y$ and the corresponding optimization problem is easier to solve than learning all mask layers simultaneously. Representations from higher layers are shown to be more context-specific than lower layer representations~\cite{ethayarajh2019contextual}. In contrast, the bottom up approach may drop key features in lower layers that significantly contribute to important higher layer features. 

We compare variants of \ourmethod by using varying backbone models and removing the corresponding components. For backbones, we compare \bart with T5 and \bert, denoted them as \textbf{\ourmethod-T5}, and \textbf{\ourmethod-\bert}. To study the contribution of each component in our approach, we conduct experiments where we exclude the mask layers, attention mechanisms, or both. These models are denoted by \textbf{w/o $\vm$}, \textbf{w/o $\va$}, and \textbf{w/o $\va\vm$}, respectively. The corresponding results are reported in Table~\ref{tab:ablation_study_sentiment_cls}, Table~\ref{tab:ablation_study_multiclass_cls} and Table~\ref{tab:ablation_study_multiclass_socialdial_cls} in Appendix. For comparison between backbones, we find that encoder-decoder neural architectures  (\ie \bart, T5) consistently achieve better performance than encoder-only models (\ie \bert). Compared with variants that remove both the attention module and mask layers, \ourmethod with the attention module or mask module has a significant performance improvement in terms of accuracy or F1 on average, which justifies the usefulness of both modules. 

Additionally, we compare \ourmethod with various sparsity methods to implement mask layers, including \textbf{STR} \cite{pmlr-v119-kusupati20a}, \textbf{STE} \cite{bengio2013estimating,LIU2020Dynamic}, and \textbf{Scalar}, which uses a learnable single scalar instead of the threshold vector $\vs$. All those alternative methods lead to a significant drop, as seen in Table~\ref{tab:ablation_study_sentiment_cls}. 

\begin{table}
\centering
\resizebox{\columnwidth}{!}{%
\begin{tabular}{l|cccc}
\hline
\textbf{}                       & \multicolumn{3}{c}{\textbf{Amazon$\rightarrow$}}   & \multicolumn{1}{l}{} \\
\textbf{Models}                 & \textbf{Yelp} & \textbf{IMDB} & \textbf{TweetEval} & \textbf{Avg.}        \\ \hline
\ourmethod-1k                   & 92.21         & 87.29         & 85.18              & 88.22                \\
\ourmethod-10k                  & 94.82         & 89.11         & 88.43              & 90.78                \\
\ourmethod-100k                 & 94.90         & 90.24         & 89.01              & 91.38                \\
\ourmethod-1M                   & 94.95         & 90.29         & 89.20              & 91.48                \\
\ourmethod-3.6M                 & 95.14         & 90.86         & 91.08              & 92.36                \\ \hline
\ourmethod - w/o $\va\vm$ -1k   & 70.62         & 68.61         & 66.07              & 68.43                \\
\ourmethod - w/o $\va\vm$ -10k  & 84.88         & 79.02         & 75.19              & 79.70                \\
\ourmethod - w/o $\va\vm$ -100k & 87.05         & 84.95         & 80.48              & 84.16                \\
\ourmethod - w/o $\va\vm$ -1M   & 91.38         & 87.06         & 81.59              & 86.68                \\
\ourmethod - w/o $\va\vm$ -3.6M & 92.20         & 88.19         & 82.69              & 87.69                \\ \hline
\end{tabular}%
}
\caption{Domain generalization experiment with different training sizes in the source domain.}
\label{tab:different_training_size}
\vspace{-15pt}
\end{table}

To explore the influence of source domain training data size on performance within target domains, we train models based on \bart with and without our method on the Amazon review dataset with varying sizes of training data (\ie 1k, 10k, 100k, 1M, and 3.6M). The results in Table~\ref{tab:different_training_size} show that our method depends significantly less on training data size, though more training data can improve the performance overall. Notably, 1k training data yields a remarkable decline for the models without using \ourmethod, while the corresponding performance reduction is significantly less by using our method. 

\section{Conclusion}
This paper presents a novel method, coined \ourmethod, which is a greedy layer-wise representation learning method aiming to improve single-source domain generalization on pre-trained deep encoders for text classification tasks. The key idea is to retain invariant features through trainable mask layers and incorporate a token-level attention module to focus on the tokens that directly lead to the prediction of labels. 
Through extensive experiments, we demonstrate that \ourmethod achieves superior OOD performance over competitive baselines on multiple datasets. The visualization of masks and attention weights empirically justifies the effectiveness of identified invariant sparse representations.

\clearpage

\section*{Limitations}
Our work focuses on the text classification task, intending to investigate how to learn invariant features to improve out-of-domain generalization. However, the proposed method has promising potential for domain generalization in various NLP tasks, such as question answering and text generation tasks. Future work may consider more tasks beyond text classification.

It is worth noting that \ourmethod needs to be trained in a large source domain. The size of the source domain should ideally exceed 10,000 samples to achieve consistently good performance. However, this requirement may pose challenges in low-resource learning scenarios. 

\section*{Ethics Statement}
This research is dedicated to augmenting the reliability and safety of text classification models, particularly in the context of domain shifts, as highlighted by \citet{ribeiro-etal-2020-beyond}. By focusing on the learning of invariant features across diverse domains, our approach aims to provide tangible benefits to applications that serve a wide array of user groups. From a user-centric perspective, the implementation of our methodology is expected to bolster the trustworthiness and diminish potential biases in language models.

It is pertinent to note that our study does not involve human subjects, nor does it contravene any legal or ethical standards. We foresee no detrimental impacts arising from our research endeavors. The experimental work underpinning this study was exclusively conducted using datasets that are publicly accessible. Our overarching goal is to foster enhanced academic and societal consciousness regarding the challenges of domain generalization in the field of natural language processing.

\bibliography{iclr2024_conference, anthology,custom}

\begin{thebibliography}{71}
\expandafter\ifx\csname natexlab\endcsname\relax\def\natexlab#1{#1}\fi

\bibitem[{Arjovsky et~al.(2020)Arjovsky, Bottou, Gulrajani, and
  Lopez-Paz}]{arjovsky2020invariant}
Martin Arjovsky, Léon Bottou, Ishaan Gulrajani, and David Lopez-Paz. 2020.
\newblock \href {http://arxiv.org/abs/1907.02893} {Invariant risk
  minimization}.

\bibitem[{Asgari et~al.(2022)Asgari, Khani, Khani, Gholami, Tran,
  Mahdavi-Amiri, and Hamarneh}]{asgari2022masktune}
Saeid Asgari, Aliasghar Khani, Fereshte Khani, Ali Gholami, Linh Tran, Ali
  Mahdavi-Amiri, and Ghassan Hamarneh. 2022.
\newblock \href {https://openreview.net/forum?id=hMGSz9PNQes} {Masktune:
  Mitigating spurious correlations by forcing to explore}.
\newblock In \emph{Advances in Neural Information Processing Systems}.

\bibitem[{Bang et~al.(2023)Bang, Cahyawijaya, Lee, Dai, Su, Wilie, Lovenia, Ji,
  Yu, Chung, Do, Xu, and Fung}]{bang2023multitask}
Yejin Bang, Samuel Cahyawijaya, Nayeon Lee, Wenliang Dai, Dan Su, Bryan Wilie,
  Holy Lovenia, Ziwei Ji, Tiezheng Yu, Willy Chung, Quyet~V. Do, Yan Xu, and
  Pascale Fung. 2023.
\newblock \href {http://arxiv.org/abs/2302.04023} {A multitask, multilingual,
  multimodal evaluation of chatgpt on reasoning, hallucination, and
  interactivity}.

\bibitem[{Barbieri et~al.(2020)Barbieri, Camacho-Collados, Espinosa~Anke, and
  Neves}]{barbieri-etal-2020-tweeteval}
Francesco Barbieri, Jose Camacho-Collados, Luis Espinosa~Anke, and Leonardo
  Neves. 2020.
\newblock \href {https://doi.org/10.18653/v1/2020.findings-emnlp.148}
  {{T}weet{E}val: Unified benchmark and comparative evaluation for tweet
  classification}.
\newblock In \emph{Findings of the Association for Computational Linguistics:
  EMNLP 2020}, pages 1644--1650, Online. Association for Computational
  Linguistics.

\bibitem[{Ben-David et~al.(2022)Ben-David, Oved, and
  Reichart}]{10.1162/tacl_a_00468}
Eyal Ben-David, Nadav Oved, and Roi Reichart. 2022.
\newblock \href {https://doi.org/10.1162/tacl_a_00468} {{PADA: Example-based
  Prompt Learning for on-the-fly Adaptation to Unseen Domains}}.
\newblock \emph{Transactions of the Association for Computational Linguistics},
  10:414--433.

\bibitem[{Bengio et~al.(2013)Bengio, Léonard, and
  Courville}]{bengio2013estimating}
Yoshua Bengio, Nicholas Léonard, and Aaron Courville. 2013.
\newblock \href {http://arxiv.org/abs/1308.3432} {Estimating or propagating
  gradients through stochastic neurons for conditional computation}.

\bibitem[{Bühlmann(2018)}]{bühlmann2018invariance}
Peter Bühlmann. 2018.
\newblock \href {http://arxiv.org/abs/1812.08233} {Invariance, causality and
  robustness}.

\bibitem[{Chattopadhyay et~al.(2020)Chattopadhyay, Balaji, and
  Hoffman}]{10.1007/978-3-030-58545-7_18}
Prithvijit Chattopadhyay, Yogesh Balaji, and Judy Hoffman. 2020.
\newblock \href {https://doi.org/https://doi.org/10.1007/978-3-030-58545-7_18}
  {Learning to balance specificity and invariance for in and out of domain
  generalization}.
\newblock In \emph{Computer Vision -- ECCV 2020}, pages 301--318, Cham.
  Springer International Publishing.

\bibitem[{Clue-AI(2023)}]{chatyuan}
Clue-AI. 2023.
\newblock Chatyuan.
\newblock \url{https://github.com/clue-ai/ChatYuan}.

\bibitem[{Del~Corso et~al.(2005)Del~Corso, Gull\'{\i}, and
  Romani}]{10.1145/1060745.1060764}
Gianna~M. Del~Corso, Antonio Gull\'{\i}, and Francesco Romani. 2005.
\newblock \href {https://doi.org/10.1145/1060745.1060764} {Ranking a stream of
  news}.
\newblock In \emph{Proceedings of the 14th International Conference on World
  Wide Web}, WWW '05, page 97–106, New York, NY, USA. Association for
  Computing Machinery.

\bibitem[{Devlin et~al.(2019)Devlin, Chang, Lee, and
  Toutanova}]{devlin-etal-2019-bert}
Jacob Devlin, Ming-Wei Chang, Kenton Lee, and Kristina Toutanova. 2019.
\newblock \href {https://doi.org/10.18653/v1/N19-1423} {{BERT}: Pre-training of
  deep bidirectional transformers for language understanding}.
\newblock In \emph{Proceedings of the 2019 Conference of the North {A}merican
  Chapter of the Association for Computational Linguistics: Human Language
  Technologies, Volume 1 (Long and Short Papers)}, pages 4171--4186,
  Minneapolis, Minnesota. Association for Computational Linguistics.

\bibitem[{Ding et~al.(2022)Ding, Wang, Liang, Liang, Wang, and
  Chen}]{ding2022domain}
Yu~Ding, Lei Wang, Bin Liang, Shuming Liang, Yang Wang, and Fang Chen. 2022.
\newblock \href {https://openreview.net/forum?id=37Rf7BTAtAM} {Domain
  generalization by learning and removing domain-specific features}.
\newblock In \emph{Advances in Neural Information Processing Systems}.

\bibitem[{Du et~al.(2022)Du, Qian, Liu, Ding, Qiu, Yang, and
  Tang}]{du-etal-2022-glm}
Zhengxiao Du, Yujie Qian, Xiao Liu, Ming Ding, Jiezhong Qiu, Zhilin Yang, and
  Jie Tang. 2022.
\newblock \href {https://doi.org/10.18653/v1/2022.acl-long.26} {{GLM}: General
  language model pretraining with autoregressive blank infilling}.
\newblock In \emph{Proceedings of the 60th Annual Meeting of the Association
  for Computational Linguistics (Volume 1: Long Papers)}, pages 320--335,
  Dublin, Ireland. Association for Computational Linguistics.

\bibitem[{Ethayarajh(2019)}]{ethayarajh2019contextual}
Kawin Ethayarajh. 2019.
\newblock How contextual are contextualized word representations? comparing the
  geometry of bert, elmo, and gpt-2 embeddings.
\newblock \emph{arXiv preprint arXiv:1909.00512}.

\bibitem[{Feng et~al.(2023)Feng, Qu, and Haffari}]{10.1162/tacl_a_00561}
Tao Feng, Lizhen Qu, and Gholamreza Haffari. 2023.
\newblock \href {https://doi.org/10.1162/tacl_a_00561} {{Less is More: Mitigate
  Spurious Correlations for Open-Domain Dialogue Response Generation Models by
  Causal Discovery}}.
\newblock \emph{Transactions of the Association for Computational Linguistics},
  11:511--530.

\bibitem[{Gokhale et~al.(2022)Gokhale, Mishra, Luo, Sachdeva, and
  Baral}]{gokhale-etal-2022-generalized}
Tejas Gokhale, Swaroop Mishra, Man Luo, Bhavdeep Sachdeva, and Chitta Baral.
  2022.
\newblock \href {https://doi.org/10.18653/v1/2022.findings-acl.213}
  {\textit{Generalized but not Robust?} comparing the effects of data
  modification methods on out-of-domain generalization and adversarial
  robustness}.
\newblock In \emph{Findings of the Association for Computational Linguistics:
  ACL 2022}, pages 2705--2718, Dublin, Ireland. Association for Computational
  Linguistics.

\bibitem[{Gulli(2005)}]{10.1145/1062745.1062778}
A.~Gulli. 2005.
\newblock \href {https://doi.org/10.1145/1062745.1062778} {The anatomy of a
  news search engine}.
\newblock In \emph{Special Interest Tracks and Posters of the 14th
  International Conference on World Wide Web}, WWW '05, page 880–881, New
  York, NY, USA. Association for Computing Machinery.

\bibitem[{Gulrajani and Lopez-Paz(2021)}]{gulrajani2021in}
Ishaan Gulrajani and David Lopez-Paz. 2021.
\newblock \href {https://openreview.net/forum?id=lQdXeXDoWtI} {In search of
  lost domain generalization}.
\newblock In \emph{International Conference on Learning Representations}.

\bibitem[{Hu et~al.(2022{\natexlab{a}})Hu, yelong shen, Wallis, Allen-Zhu, Li,
  Wang, Wang, and Chen}]{hu2022lora}
Edward~J Hu, yelong shen, Phillip Wallis, Zeyuan Allen-Zhu, Yuanzhi Li, Shean
  Wang, Lu~Wang, and Weizhu Chen. 2022{\natexlab{a}}.
\newblock \href {https://openreview.net/forum?id=nZeVKeeFYf9} {Lo{RA}: Low-rank
  adaptation of large language models}.
\newblock In \emph{International Conference on Learning Representations}.

\bibitem[{Hu et~al.(2022{\natexlab{b}})Hu, Zhao, Yi, Yao, Hong, Sun, and
  Chi}]{NEURIPS2022_4a9eaf6d}
Ziniu Hu, Zhe Zhao, Xinyang Yi, Tiansheng Yao, Lichan Hong, Yizhou Sun, and
  Ed~Chi. 2022{\natexlab{b}}.
\newblock \href
  {https://proceedings.neurips.cc/paper_files/paper/2022/file/4a9eaf6dff3fdac9ab1aaf4c0fe2d563-Paper-Conference.pdf}
  {Improving multi-task generalization via regularizing spurious correlation}.
\newblock In \emph{Advances in Neural Information Processing Systems},
  volume~35, pages 11450--11466. Curran Associates, Inc.

\bibitem[{Izmailov et~al.(2022)Izmailov, Kirichenko, Gruver, and
  Wilson}]{NEURIPS2022_fb64a552}
Pavel Izmailov, Polina Kirichenko, Nate Gruver, and Andrew~G Wilson. 2022.
\newblock \href
  {https://proceedings.neurips.cc/paper_files/paper/2022/file/fb64a552feda3d981dbe43527a80a07e-Paper-Conference.pdf}
  {On feature learning in the presence of spurious correlations}.
\newblock In \emph{Advances in Neural Information Processing Systems},
  volume~35, pages 38516--38532. Curran Associates, Inc.

\bibitem[{Jia and Zhang(2022)}]{jia-zhang-2022-prompt}
Chen Jia and Yue Zhang. 2022.
\newblock \href {https://aclanthology.org/2022.emnlp-main.690} {Prompt-based
  distribution alignment for domain generalization in text classification}.
\newblock In \emph{Proceedings of the 2022 Conference on Empirical Methods in
  Natural Language Processing}, pages 10147--10157, Abu Dhabi, United Arab
  Emirates. Association for Computational Linguistics.

\bibitem[{Kingma and Ba(2015)}]{kingma2017adam}
Diederik~P. Kingma and Jimmy Ba. 2015.
\newblock \href {http://arxiv.org/abs/1412.6980} {Adam: {A} method for
  stochastic optimization}.

\bibitem[{Kusupati et~al.(2020)Kusupati, Ramanujan, Somani, Wortsman, Jain,
  Kakade, and Farhadi}]{pmlr-v119-kusupati20a}
Aditya Kusupati, Vivek Ramanujan, Raghav Somani, Mitchell Wortsman, Prateek
  Jain, Sham Kakade, and Ali Farhadi. 2020.
\newblock \href {https://proceedings.mlr.press/v119/kusupati20a.html} {Soft
  threshold weight reparameterization for learnable sparsity}.
\newblock In \emph{Proceedings of the 37th International Conference on Machine
  Learning}, volume 119 of \emph{Proceedings of Machine Learning Research},
  pages 5544--5555. PMLR.

\bibitem[{Lewis et~al.(2020)Lewis, Liu, Goyal, Ghazvininejad, Mohamed, Levy,
  Stoyanov, and Zettlemoyer}]{lewis-etal-2020-bart}
Mike Lewis, Yinhan Liu, Naman Goyal, Marjan Ghazvininejad, Abdelrahman Mohamed,
  Omer Levy, Veselin Stoyanov, and Luke Zettlemoyer. 2020.
\newblock \href {https://doi.org/10.18653/v1/2020.acl-main.703} {{BART}:
  Denoising sequence-to-sequence pre-training for natural language generation,
  translation, and comprehension}.
\newblock In \emph{Proceedings of the 58th Annual Meeting of the Association
  for Computational Linguistics}, pages 7871--7880, Online. Association for
  Computational Linguistics.

\bibitem[{Li et~al.(2019)Li, Zhang, Gan, Cheng, Brockett, Dolan, and
  Sun}]{li-etal-2019-domain}
Dianqi Li, Yizhe Zhang, Zhe Gan, Yu~Cheng, Chris Brockett, Bill Dolan, and
  Ming-Ting Sun. 2019.
\newblock \href {https://doi.org/10.18653/v1/D19-1325} {Domain adaptive text
  style transfer}.
\newblock In \emph{Proceedings of the 2019 Conference on Empirical Methods in
  Natural Language Processing and the 9th International Joint Conference on
  Natural Language Processing (EMNLP-IJCNLP)}, pages 3304--3313, Hong Kong,
  China. Association for Computational Linguistics.

\bibitem[{Li et~al.(2018{\natexlab{a}})Li, Pan, Wang, and Kot}]{8578664}
Haoliang Li, Sinno~Jialin Pan, Shiqi Wang, and Alex~C. Kot. 2018{\natexlab{a}}.
\newblock \href {https://doi.org/10.1109/CVPR.2018.00566} {Domain
  generalization with adversarial feature learning}.
\newblock In \emph{2018 IEEE/CVF Conference on Computer Vision and Pattern
  Recognition}, pages 5400--5409.

\bibitem[{Li et~al.(2018{\natexlab{b}})Li, Tian, Gong, Liu, Liu, Zhang, and
  Tao}]{10.1007/978-3-030-01267-0_38}
Ya~Li, Xinmei Tian, Mingming Gong, Yajing Liu, Tongliang Liu, Kun Zhang, and
  Dacheng Tao. 2018{\natexlab{b}}.
\newblock \href {https://doi.org/https://doi.org/10.1007/978-3-030-01267-0_38}
  {Deep domain generalization via conditional invariant adversarial networks}.
\newblock In \emph{Computer Vision -- ECCV 2018}, pages 647--663, Cham.
  Springer International Publishing.

\bibitem[{Liu et~al.(2020)Liu, XU, SHI, Cheung, and So}]{LIU2020Dynamic}
Junjie Liu, Zhe XU, Runbin SHI, Ray C.~C. Cheung, and Hayden~K.H. So. 2020.
\newblock \href {https://openreview.net/forum?id=SJlbGJrtDB} {Dynamic sparse
  training: Find efficient sparse network from scratch with trainable masked
  layers}.
\newblock In \emph{International Conference on Learning Representations}.

\bibitem[{Liu et~al.(2022)Liu, Yoo, Xing, Oh, Fakhri, Kang, and
  Woo}]{liu2022deepUDAReview}
Xiaofeng Liu, Chaehwa Yoo, Fangxu Xing, Hyejin Oh, Georges~El Fakhri, Je-Won
  Kang, and Jonghye Woo. 2022.
\newblock \href {https://doi.org/10.1561/116.00000192} {Deep unsupervised
  domain adaptation: A review of recent advances and perspectives}.
\newblock \emph{APSIPA Transactions on Signal and Information Processing},
  11(1).

\bibitem[{Lv et~al.(2022)Lv, Liang, Li, Zang, Liu, Wang, and
  Liu}]{lv2022causality}
Fangrui Lv, Jian Liang, Shuang Li, Bin Zang, Chi~Harold Liu, Ziteng Wang, and
  Di~Liu. 2022.
\newblock \href {https://doi.org/10.1109/CVPR52688.2022.00788} {Causality
  inspired representation learning for domain generalization}.
\newblock In \emph{2022 IEEE/CVF Conference on Computer Vision and Pattern
  Recognition (CVPR)}, pages 8036--8046, Los Alamitos, CA, USA. IEEE Computer
  Society.

\bibitem[{Maas et~al.(2011)Maas, Daly, Pham, Huang, Ng, and
  Potts}]{maas-EtAl:2011:ACL-HLT2011}
Andrew~L. Maas, Raymond~E. Daly, Peter~T. Pham, Dan Huang, Andrew~Y. Ng, and
  Christopher Potts. 2011.
\newblock \href {http://www.aclweb.org/anthology/P11-1015} {Learning word
  vectors for sentiment analysis}.
\newblock In \emph{Proceedings of the 49th Annual Meeting of the Association
  for Computational Linguistics: Human Language Technologies}, pages 142--150,
  Portland, Oregon, USA. Association for Computational Linguistics.

\bibitem[{Min et~al.(2022)Min, Lyu, Holtzman, Artetxe, Lewis, Hajishirzi, and
  Zettlemoyer}]{min-etal-2022-rethinking}
Sewon Min, Xinxi Lyu, Ari Holtzman, Mikel Artetxe, Mike Lewis, Hannaneh
  Hajishirzi, and Luke Zettlemoyer. 2022.
\newblock \href {https://aclanthology.org/2022.emnlp-main.759} {Rethinking the
  role of demonstrations: What makes in-context learning work?}
\newblock In \emph{Proceedings of the 2022 Conference on Empirical Methods in
  Natural Language Processing}, pages 11048--11064, Abu Dhabi, United Arab
  Emirates. Association for Computational Linguistics.

\bibitem[{Muandet et~al.(2013)Muandet, Balduzzi, and
  Sch\"{o}lkopf}]{10.5555/3042817.3042820}
Krikamol Muandet, David Balduzzi, and Bernhard Sch\"{o}lkopf. 2013.
\newblock \href {https://dl.acm.org/doi/10.5555/3042817.3042820} {Domain
  generalization via invariant feature representation}.
\newblock In \emph{Proceedings of the 30th International Conference on
  International Conference on Machine Learning - Volume 28}, ICML'13, page
  I–10–I–18. JMLR.org.

\bibitem[{Neal(2020)}]{neal2020introduction}
Brady Neal. 2020.
\newblock Introduction to causal inference.
\newblock \emph{Course Lecture Notes (draft)}.

\bibitem[{OpenAI(2023)}]{openai2023gpt4}
OpenAI. 2023.
\newblock \href {http://arxiv.org/abs/2303.08774} {Gpt-4 technical report}.

\bibitem[{Ouyang et~al.(2023)Ouyang, Chen, Li, Li, Qin, Bai, and
  Rueckert}]{9961940}
Cheng Ouyang, Chen Chen, Surui Li, Zeju Li, Chen Qin, Wenjia Bai, and Daniel
  Rueckert. 2023.
\newblock \href {https://doi.org/10.1109/TMI.2022.3224067} {Causality-inspired
  single-source domain generalization for medical image segmentation}.
\newblock \emph{IEEE Transactions on Medical Imaging}, 42(4):1095--1106.

\bibitem[{Peters et~al.(2016)Peters, Bühlmann, and
  Meinshausen}]{10.2307/44682904}
Jonas Peters, Peter Bühlmann, and Nicolai Meinshausen. 2016.
\newblock \href {http://www.jstor.org/stable/44682904} {Causal inference by
  using invariant prediction: identification and confidence intervals}.
\newblock \emph{Journal of the Royal Statistical Society. Series B (Statistical
  Methodology)}, 78(5):947--1012.

\bibitem[{Peters et~al.(2017)Peters, Janzing, and Schlkopf}]{10.5555/3202377}
Jonas Peters, Dominik Janzing, and Bernhard Schlkopf. 2017.
\newblock \href {https://dl.acm.org/doi/10.5555/3202377} {\emph{Elements of
  Causal Inference: Foundations and Learning Algorithms}}.
\newblock The MIT Press.

\bibitem[{Qiao et~al.(2020)Qiao, Zhao, and Peng}]{qiao2020learning}
Fengchun Qiao, Long Zhao, and Xi~Peng. 2020.
\newblock \href
  {https://openaccess.thecvf.com/content_CVPR_2020/html/Qiao_Learning_to_Learn_Single_Domain_Generalization_CVPR_2020_paper.html}
  {Learning to learn single domain generalization}.
\newblock In \emph{Proceedings of the IEEE/CVF Conference on Computer Vision
  and Pattern Recognition}, pages 12556--12565.

\bibitem[{Quinzan et~al.(2023)Quinzan, Soleymani, Jaillet, Rojas, and
  Bauer}]{quinzan2023drcfs}
Francesco Quinzan, Ashkan Soleymani, Patrick Jaillet, Cristian~R Rojas, and
  Stefan Bauer. 2023.
\newblock Drcfs: Doubly robust causal feature selection.
\newblock In \emph{International Conference on Machine Learning}, pages
  28468--28491. PMLR.

\bibitem[{Raffel et~al.(2020)Raffel, Shazeer, Roberts, Lee, Narang, Matena,
  Zhou, Li, and Liu}]{10.5555/3455716.3455856}
Colin Raffel, Noam Shazeer, Adam Roberts, Katherine Lee, Sharan Narang, Michael
  Matena, Yanqi Zhou, Wei Li, and Peter~J. Liu. 2020.
\newblock \href {https://dl.acm.org/doi/abs/10.5555/3455716.3455856} {Exploring
  the limits of transfer learning with a unified text-to-text transformer}.
\newblock \emph{J. Mach. Learn. Res.}, 21(1).

\bibitem[{Ribeiro et~al.(2020)Ribeiro, Wu, Guestrin, and
  Singh}]{ribeiro-etal-2020-beyond}
Marco~Tulio Ribeiro, Tongshuang Wu, Carlos Guestrin, and Sameer Singh. 2020.
\newblock \href {https://doi.org/10.18653/v1/2020.acl-main.442} {Beyond
  accuracy: Behavioral testing of {NLP} models with {C}heck{L}ist}.
\newblock In \emph{Proceedings of the 58th Annual Meeting of the Association
  for Computational Linguistics}, pages 4902--4912, Online. Association for
  Computational Linguistics.

\bibitem[{Saunders(2022)}]{saunders2022DASurveyMT}
Danielle Saunders. 2022.
\newblock \href {https://doi.org/10.1613/jair.1.13566} {Domain adaptation and
  multi-domain adaptation for neural machine translation: A survey}.
\newblock \emph{Journal of Artificial Intelligence Research}, 75.

\bibitem[{Shao et~al.(2019)Shao, Lan, Li, and Yuen}]{8953226}
Rui Shao, Xiangyuan Lan, Jiawei Li, and Pong~C. Yuen. 2019.
\newblock \href {https://doi.org/10.1109/CVPR.2019.01026} {Multi-adversarial
  discriminative deep domain generalization for face presentation attack
  detection}.
\newblock In \emph{2019 IEEE/CVF Conference on Computer Vision and Pattern
  Recognition (CVPR)}, pages 10015--10023.

\bibitem[{Shao et~al.(2021)Shao, Geng, Liu, Dai, Yan, Yang, Zhe, Bao, and
  Qiu}]{shao2021cpt}
Yunfan Shao, Zhichao Geng, Yitao Liu, Junqi Dai, Hang Yan, Fei Yang, Li~Zhe,
  Hujun Bao, and Xipeng Qiu. 2021.
\newblock \href {https://arxiv.org/abs/2109.05729} {Cpt: A pre-trained
  unbalanced transformer for both chinese language understanding and
  generation}.
\newblock \emph{arXiv preprint arXiv:2109.05729}.

\bibitem[{Shen et~al.(2021)Shen, Liu, He, Zhang, Xu, Yu, and
  Cui}]{shen2021outofdistribution}
Zheyan Shen, Jiashuo Liu, Yue He, Xingxuan Zhang, Renzhe Xu, Han Yu, and Peng
  Cui. 2021.
\newblock \href {http://arxiv.org/abs/2108.13624} {Towards out-of-distribution
  generalization: A survey}.

\bibitem[{Shiri et~al.(2023)Shiri, Zhuo, Li, Nguyen, Pan, Wang, Haffari, and
  Li}]{shiri2023paraphrasing}
Fatemeh Shiri, Terry~Yue Zhuo, Zhuang Li, Van Nguyen, Shirui Pan, Weiqing Wang,
  Reza Haffari, and Yuan-Fang Li. 2023.
\newblock \href {http://arxiv.org/abs/2203.10854} {Paraphrasing techniques for
  maritime qa system}.

\bibitem[{Taori et~al.(2023)Taori, Gulrajani, Zhang, Dubois, Li, Guestrin,
  Liang, and Hashimoto}]{alpaca}
Rohan Taori, Ishaan Gulrajani, Tianyi Zhang, Yann Dubois, Xuechen Li, Carlos
  Guestrin, Percy Liang, and Tatsunori~B. Hashimoto. 2023.
\newblock Stanford alpaca: An instruction-following llama model.
\newblock \url{https://github.com/tatsu-lab/stanford_alpaca}.

\bibitem[{THUDM(2023)}]{chatglm_6b}
THUDM. 2023.
\newblock Chatglm-6b.
\newblock \url{https://github.com/THUDM/ChatGLM-6B}.

\bibitem[{Touvron et~al.(2023)Touvron, Lavril, Izacard, Martinet, Lachaux,
  Lacroix, Rozière, Goyal, Hambro, Azhar, Rodriguez, Joulin, Grave, and
  Lample}]{touvron2023llama}
Hugo Touvron, Thibaut Lavril, Gautier Izacard, Xavier Martinet, Marie-Anne
  Lachaux, Timothée Lacroix, Baptiste Rozière, Naman Goyal, Eric Hambro,
  Faisal Azhar, Aurelien Rodriguez, Armand Joulin, Edouard Grave, and Guillaume
  Lample. 2023.
\newblock \href {http://arxiv.org/abs/2302.13971} {Llama: Open and efficient
  foundation language models}.

\bibitem[{Veitch et~al.(2021)Veitch, D'Amour, Yadlowsky, and
  Eisenstein}]{veitch2021counterfactual}
Victor Veitch, Alexander D'Amour, Steve Yadlowsky, and Jacob Eisenstein. 2021.
\newblock Counterfactual invariance to spurious correlations: Why and how to
  pass stress tests.
\newblock \emph{arXiv preprint arXiv:2106.00545}.

\bibitem[{Volpi and Murino(2019)}]{volpi2019addressing}
Riccardo Volpi and Vittorio Murino. 2019.
\newblock \href
  {https://openaccess.thecvf.com/content_ICCV_2019/html/Volpi_Addressing_Model_Vulnerability_to_Distributional_Shifts_Over_Image_Transformation_Sets_ICCV_2019_paper.html}
  {Addressing model vulnerability to distributional shifts over image
  transformation sets}.
\newblock In \emph{Proceedings of the IEEE/CVF International Conference on
  Computer Vision}, pages 7980--7989.

\bibitem[{Wang(2023)}]{alpaca-lora}
Eric~J. Wang. 2023.
\newblock Alpaca-lora.
\newblock \url{https://github.com/tloen/alpaca-lora}.

\bibitem[{Wang et~al.(2021{\natexlab{a}})Wang, Huang, and
  Xing}]{wang2020CounteringSpuriousCorrelations}
Haohan Wang, Zeyi Huang, and Eric Xing. 2021{\natexlab{a}}.
\newblock \href {https://openreview.net/forum?id=o21sjfFaU1} {Learning robust
  models by countering spurious correlations}.

\bibitem[{Wang et~al.(2021{\natexlab{b}})Wang, Lan, Liu, Ouyang, and
  Qin}]{wang2022surveyDomainGeneralizing}
Jindong Wang, Cuiling Lan, Chang Liu, Yidong Ouyang, and Tao Qin.
  2021{\natexlab{b}}.
\newblock \href {https://doi.org/10.24963/ijcai.2021/628} {Generalizing to
  unseen domains: A survey on domain generalization}.
\newblock pages 4627--4635.
\newblock Survey Track.

\bibitem[{Wang and Jordan(2022)}]{wang2022representation}
Yixin Wang and Michael Jordan. 2022.
\newblock \href {https://openreview.net/forum?id=_rqQZ_HL_fY} {Representation
  learning as finding necessary and sufficient causes}.
\newblock In \emph{ICML 2022: Workshop on Spurious Correlations, Invariance and
  Stability}.

\bibitem[{Wang et~al.(2022)Wang, Kordi, Mishra, Liu, Smith, Khashabi, and
  Hajishirzi}]{wang2022selfinstruct}
Yizhong Wang, Yeganeh Kordi, Swaroop Mishra, Alisa Liu, Noah~A. Smith, Daniel
  Khashabi, and Hannaneh Hajishirzi. 2022.
\newblock \href {http://arxiv.org/abs/2212.10560} {Self-instruct: Aligning
  language model with self generated instructions}.

\bibitem[{Wang et~al.(2023)Wang, Xie, Ding, Feng, and Xia}]{wang2023chatgpt}
Zengzhi Wang, Qiming Xie, Zixiang Ding, Yi~Feng, and Rui Xia. 2023.
\newblock \href {http://arxiv.org/abs/2304.04339} {Is chatgpt a good sentiment
  analyzer? a preliminary study}.

\bibitem[{Wang et~al.(2021{\natexlab{c}})Wang, Luo, Qiu, Huang, and
  Baktashmotlagh}]{wang2021learning}
Zijian Wang, Yadan Luo, Ruihong Qiu, Zi~Huang, and Mahsa Baktashmotlagh.
  2021{\natexlab{c}}.
\newblock \href
  {https://openaccess.thecvf.com/content/ICCV2021/html/Wang_Learning_To_Diversify_for_Single_Domain_Generalization_ICCV_2021_paper.html}
  {Learning to diversify for single domain generalization}.
\newblock In \emph{Proceedings of the IEEE/CVF International Conference on
  Computer Vision}, pages 834--843.

\bibitem[{Wei and Zou(2019)}]{wei-zou-2019-eda}
Jason Wei and Kai Zou. 2019.
\newblock \href {https://doi.org/10.18653/v1/D19-1670} {{EDA}: Easy data
  augmentation techniques for boosting performance on text classification
  tasks}.
\newblock In \emph{Proceedings of the 2019 Conference on Empirical Methods in
  Natural Language Processing and the 9th International Joint Conference on
  Natural Language Processing (EMNLP-IJCNLP)}, pages 6382--6388, Hong Kong,
  China. Association for Computational Linguistics.

\bibitem[{Wiles et~al.(2022)Wiles, Gowal, Stimberg, Rebuffi, Ktena, Dvijotham,
  and Cemgil}]{wiles2022a}
Olivia Wiles, Sven Gowal, Florian Stimberg, Sylvestre-Alvise Rebuffi, Ira
  Ktena, Krishnamurthy~Dj Dvijotham, and Ali~Taylan Cemgil. 2022.
\newblock \href {https://openreview.net/forum?id=Dl4LetuLdyK} {A fine-grained
  analysis on distribution shift}.
\newblock In \emph{International Conference on Learning Representations}.

\bibitem[{Xie et~al.(2020)Xie, Dai, Hovy, Luong, and
  Le}]{10.5555/3495724.3496249}
Qizhe Xie, Zihang Dai, Eduard Hovy, Minh-Thang Luong, and Quoc~V. Le. 2020.
\newblock \href {https://dl.acm.org/doi/10.5555/3495724.3496249} {Unsupervised
  data augmentation for consistency training}.
\newblock In \emph{Proceedings of the 34th International Conference on Neural
  Information Processing Systems}, NIPS'20, Red Hook, NY, USA. Curran
  Associates Inc.

\bibitem[{Xu and Cheung(2019)}]{DBLP:conf/bmvc/XuC19}
Zhe Xu and Ray C.~C. Cheung. 2019.
\newblock \href {https://bmvc2019.org/wp-content/uploads/papers/0588-paper.pdf}
  {Accurate and compact convolutional neural networks with trained
  binarization}.
\newblock In \emph{30th British Machine Vision Conference 2019, {BMVC} 2019,
  Cardiff, UK, September 9-12, 2019}, page~19. {BMVA} Press.

\bibitem[{Yang et~al.(2023)Yang, Jin, Tang, Han, Feng, Jiang, Yin, and
  Hu}]{yang2023harnessing}
Jingfeng Yang, Hongye Jin, Ruixiang Tang, Xiaotian Han, Qizhang Feng, Haoming
  Jiang, Bing Yin, and Xia Hu. 2023.
\newblock \href {http://arxiv.org/abs/2304.13712} {Harnessing the power of llms
  in practice: A survey on chatgpt and beyond}.

\bibitem[{Zhan et~al.(2023)Zhan, Li, Wang, Luo, Feng, Kang, Hua, Qu, Soon,
  Sharma, Zukerman, Semnani-Azad, and Haffari}]{zhan2023socialdial}
Haolan Zhan, Zhuang Li, Yufei Wang, Linhao Luo, Tao Feng, Xiaoxi Kang, Yuncheng
  Hua, Lizhen Qu, Lay-Ki Soon, Suraj Sharma, Ingrid Zukerman, Zhaleh
  Semnani-Azad, and Gholamreza Haffari. 2023.
\newblock \href {http://arxiv.org/abs/2304.12026} {Socialdial: A benchmark for
  socially-aware dialogue systems}.

\bibitem[{Zhang et~al.(2021)Zhang, Ahuja, Xu, Wang, and
  Courville}]{zhang2022SubnetworkOOD}
Dinghuai Zhang, Kartik Ahuja, Yilun Xu, Yisen Wang, and Aaron Courville. 2021.
\newblock \href {http://proceedings.mlr.press/v139/zhang21a/zhang21a.pdf} {Can
  subnetwork structure be the key to out-of-distribution generalization?}
\newblock In \emph{International Conference on Machine Learning}, pages
  12356--12367. PMLR.

\bibitem[{Zhang et~al.(2022)Zhang, Zhang, Liu, Weller, Schölkopf, and
  Xing}]{zhang2022towards}
Hanlin Zhang, Yi-Fan Zhang, Weiyang Liu, Adrian Weller, Bernhard Schölkopf,
  and Eric~P. Xing. 2022.
\newblock \href {https://doi.org/10.1109/CVPR52688.2022.00786} {Towards
  principled disentanglement for domain generalization}.
\newblock In \emph{2022 IEEE/CVF Conference on Computer Vision and Pattern
  Recognition (CVPR)}, pages 8014--8024.

\bibitem[{Zhang et~al.(2015{\natexlab{a}})Zhang, Zhao, and
  LeCun}]{zhang2015character}
Xiang Zhang, Junbo Zhao, and Yann LeCun. 2015{\natexlab{a}}.
\newblock \href {https://dl.acm.org/doi/10.5555/2969239.2969312}
  {Character-level convolutional networks for text classification}.
\newblock page 649–657.

\bibitem[{Zhang et~al.(2015{\natexlab{b}})Zhang, Zhao, and
  LeCun}]{NIPS2015_250cf8b5}
Xiang Zhang, Junbo Zhao, and Yann LeCun. 2015{\natexlab{b}}.
\newblock \href
  {https://proceedings.neurips.cc/paper_files/paper/2015/file/250cf8b51c773f3f8dc8b4be867a9a02-Paper.pdf}
  {Character-level convolutional networks for text classification}.
\newblock In \emph{Advances in Neural Information Processing Systems},
  volume~28. Curran Associates, Inc.

\bibitem[{Zhao et~al.(2020)Zhao, Liu, Peng, and
  Metaxas}]{10.5555/3495724.3496934}
Long Zhao, Ting Liu, Xi~Peng, and Dimitris Metaxas. 2020.
\newblock \href {https://dl.acm.org/doi/abs/10.5555/3495724.3496934}
  {Maximum-entropy adversarial data augmentation for improved generalization
  and robustness}.
\newblock In \emph{Proceedings of the 34th International Conference on Neural
  Information Processing Systems}, NIPS'20, Red Hook, NY, USA. Curran
  Associates Inc.

\end{thebibliography}

\appendix
\section{Appendix}
\subsection{Background of Causal Representation Learning}
Causal representation learning aims to learn and leverage the causal relations within data to enhance model generalization and robustness against distribution shifts in the data generation process \cite{shen2021outofdistribution}. This approach differs from traditional machine learning methods that predominantly focus on correlational patterns without distinguishing causation and correlation. Causation refers to the underlying mechanisms that connect variables, implying that alterations in a causal variable will consequentially affect the associated effect variable, a process known as an intervention. In contrast, correlation does not necessarily indicate a direct mechanistic link. For instance, a model might infer 'it is raining' upon observing people with open umbrellas, which recognizes a correlation between these events. However, the act of closing umbrellas does not influence the weather. This example shows the difference between correlation and causation. Predictions relying on correlation may yield erroneous outcomes when the environment (\ie data distributions) change. For example, if umbrellas are opened due to sunlight rather than rain, a model trained on the correlation might inaccurately predict rain. This indicates the importance of making predictions based on causation rather than correlation. A causal model should predict the weather based on temperature, humidity, air pressure, etc. Prediction based on causation can enhance out-of-distribution performance, which is supported by the assumption that causal relations remain constant across diverse environments \cite{bühlmann2018invariance}. 

However, learning the complete causal structure is ambitious and may not be realized in practice. A more feasible approach involves identifying invariant features that reliably predict target variables across varying environments. A series of methods \cite{10.5555/3042817.3042820, 10.1007/978-3-030-58545-7_18, asgari2022masktune, NEURIPS2022_fb64a552, NEURIPS2022_4a9eaf6d} have been proposed by leveraging the invariance between environments. They leverage the fact that when conditioning all direct causes of a target variable, the conditional distribution of the target will not change when interventions are applied to all other variables in the model except the target itself. Building upon this foundational idea, our work seeks to identify and utilize these direct causes (\ie invariant features across environments) for the accurate prediction of target variables in the out-of-distribution setting.

\subsection{Experiment Datasets}
\label{apx:experiment_datasets}
AG News~\cite{10.1145/1062745.1062778, 10.1145/1060745.1060764, NIPS2015_250cf8b5} is a collection of news articles used for topic classification, which contains news titles, and news descriptions assigned to four topic classes. Titles and descriptions are employed as different domains. 
For social factor prediction, we use SocialDial \cite{zhan2023socialdial}, which is a Chinese socially-aware dialogue corpus consisting of synthetic conversations generated by \chatgpt and human-written conversations. Both are annotated with social factors such as location, social distance, and social relation. Synthetic conversations and human-written conversations are considered as different domains. The statistics of datasets are listed in Table~\ref{tab:dataset_statistics}. \begin{table}[h]
\centering
\resizebox{\columnwidth}{!}{%
\begin{tabular}{llccc}
\hline
\multicolumn{5}{c}{\textbf{Binary Classification}}                                                           \\
\textbf{Dataset} & \textbf{Domain}                     & \textbf{\#Train} & \textbf{\#Dev} & \textbf{\#Test} \\ \hline
Amazon           & Review of products                  & 3.6M             & 0              & 40k             \\
IMDB             & Review of movies                    & 25k              & 0              & 25k             \\
Yelp             & Review of businesses                & 560k             & 0              & 38k             \\
TweetEval        & Tweet                               & 25k              & 1k             & 6k              \\
Yahoo            & Questions from Yahoo! Answers       & 4k               & 2k             & 1k              \\ \hline
\multicolumn{5}{l}{}                                                                                         \\ \hline
\multicolumn{5}{c}{\textbf{Multi-class Classification}}                                                      \\
\textbf{Dataset} & \textbf{Domain}                     & \textbf{\#Train} & \textbf{\#Dev} & \textbf{\#Test} \\ \hline
AG News          & Title of news articles              & 120k             & 0              & 7k              \\
AG News          & Description of news articles        & 120k             & 0              & 7k              \\
SocialDial       & Synthetic conversations by \chatgpt & 68k              & 7k             & 7k              \\
SocialDial       & Human-written conversations         & 0                & 0              & 5k              \\ \hline
\end{tabular}%
}
\caption{Statistics of datasets.}
\label{tab:dataset_statistics}
\end{table}

\subsection{Training details}
\label{apx:training_details}
We use the encoder of \bart \cite{lewis-etal-2020-bart} as the default pre-trained language model. All models are trained up to 100 epochs with a minibatch size of 32 in the source domain.  We use the Adam \citep{kingma2017adam} optimizer with hyperparameters tuned on the validation sets. As a result, we run Adam with $\beta_{1}=0.9$ and $\beta_{2}=0.999$. The learning rate is $5\times 10^{-5}$. We use a linear learning rate scheduler that dynamically decreases the learning rate after a warm-up period.  All experiments are conducted on NVIDIA A40 GPU.

The process of model selection in domain generalization is inherently a learning problem. In this approach, we employ training-domain validation, which is one of the three selection methods introduced by~\citet{gulrajani2021in}. We divide each training domain into separate training and validation sets. Models are trained on the training set, and the model that achieves the highest accuracy on the validation set is chosen as the selected model.

When using large language models to predict target classification labels, the query template for sentiment analysis is: ``There are some examples about sentiment analysis: \{examples\}. Given text: \{sentence\}, what is the sentiment conveyed? Please select the answer from `positive' or `negative'.''. The query template for AG News topic classification is "There are some examples for topic classification: \{examples\}. Given text: \{sentence\}, what is the topic of this text? Please select the answer from `Business', `Sci/Tech', `World' or `Sports'.'' The query templates for SocialDial are ``There are some examples for classification: \{examples\}. Given conversation: \{conversation\}, what's the location/social distance/social relation of this conversation? Please select the answer from \{labels\}''\footnote{Since SocialDial is a dataset in Chinese, we provided queries translated from Chinese to English for use with \chatgpt.} \cite{min-etal-2022-rethinking, wang2023chatgpt, yang2023harnessing}.  

\subsection{Visual Explanation}
\begin{figure}[t]
    \centering
    \begin{subfigure}{.45\columnwidth}
        \centering
        \includegraphics[width=\columnwidth]{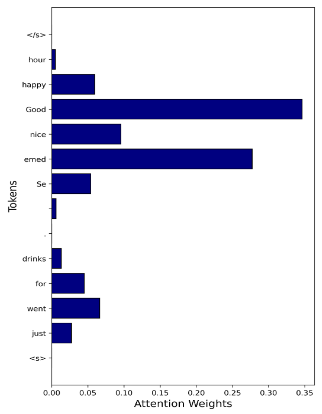}
        \caption{The sentiment label is positive.}
    \end{subfigure} \hfill
    \begin{subfigure}{.45\columnwidth}
        \centering
        \includegraphics[width=\columnwidth]{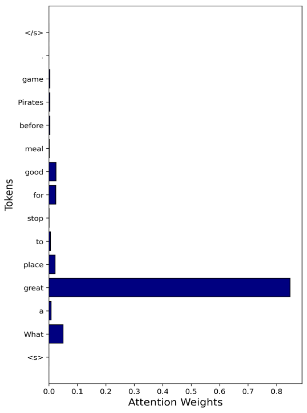}
        \caption{The sentiment label is positive.}
    \end{subfigure}
    \begin{subfigure}{.45\columnwidth}
        \centering
        \includegraphics[width=\columnwidth]{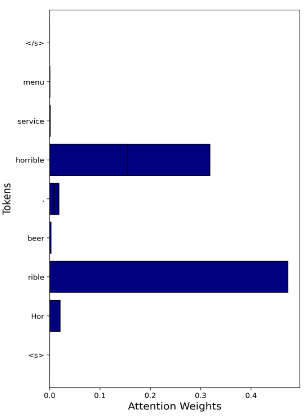}
        \caption{The sentiment label is negative.}
    \end{subfigure} \hfill
    \begin{subfigure}{.45\columnwidth}
        \centering
        \includegraphics[width=\columnwidth]{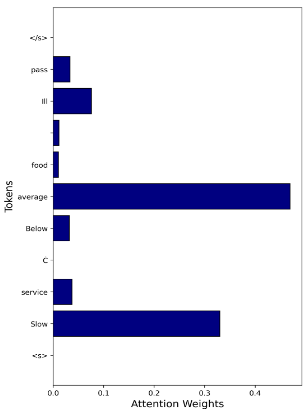}
        \caption{The sentiment label is negative.}
    \end{subfigure}
    \caption{Visualization of attention weights on tokens in Yelp dataset reviews.}
    \label{fig:attention_weights_yelp}
\end{figure}

To intuitively show how the attention module and mask module work in models, we visualize attention weights on tokens and mask vectors in Figure~\ref{fig:attention_weights_yelp} and~\ref{fig:mask_layer_visual_bart}, respectively. We also demonstrate cosine similarities between mask vectors $\vm$ trained on different source domains and Jaccard similarities between binary vectors $\vq$ trained on different source domains on Table~\ref{tab:mask_vector_cosine_similarity} and Table~\ref{tab:binary_vector_Jaccard_similarity}, respectively.

From Figure~\ref{fig:attention_weights_yelp}, we can find that our model primarily focuses its attention on sentiment-indicative tokens. Notably, positive reviews exhibit high attention weights for tokens like `good,' `great,' and `nice,' indicating their significance. Conversely, negative reviews assign high attention weights to tokens such as `horrible' and `slow,' highlighting their importance in expressing negativity.

In Figure~\ref{fig:mask_layer_visual_bart}, we visualize mask vectors $\vm$ and binary vectors $\vq$ trained on different source domains across dimensions. It can be observed that certain dimensions are consistently assigned zero (or non-zero) values across different training domains, indicating our mask layers can capture some features that are irrelevant (or invariant) across domains. We quantify invariant features across domains by computing vector similarity. We calculate cosine similarities between different mask vectors $\vm$. The results are shown in Table~\ref{tab:mask_vector_cosine_similarity}. We can find that most mask vector pairs have over $0.75$ similarity, except the Yelp-TweetEval pair, which is probably because of a larger divergence between Yelp and TweetEval domains. Table~\ref{tab:binary_vector_Jaccard_similarity} shows Jaccard similarities between binary vectors $\vq$. Most binary vector pairs have similarities of over $0.5$, except the Yelp-TweetEval pair, with a similarity of $0.45$.

\begin{table}[t]
\centering
\resizebox{\columnwidth}{!}{%
\begin{tabular}{l|cccc}
\hline
                   & \textbf{Yelp} & \textbf{Amazon} & \textbf{IMDB} & \textbf{TweetEval} \\ \hline
\textbf{Yelp}      & 1.0                               & 0.7930                              & 0.7533                            & 0.6838                                 \\
\textbf{Amazon}    & -                                 & 1.0                                 & 0.8458                            & 0.7687                                 \\
\textbf{IMDB}      & -                                 & -                                   & 1.0                               & 0.8069                                 \\
\textbf{TweetEval} & -                                 & -                                   & -                                 & 1.0                                    \\ \hline
\end{tabular}%
}
\caption{Cosine similarities between mask vectors $\vm$ trained on different source domains.}
\label{tab:mask_vector_cosine_similarity}
\end{table}

\begin{table}[t]
\centering
\resizebox{\columnwidth}{!}{%
\begin{tabular}{l|cccc}
\hline
                   & \textbf{Yelp} & \textbf{Amazon} & \textbf{IMDB} & \textbf{TweetEval} \\ \hline
\textbf{Yelp}      & 1.0           & 0.5869          & 0.5231        & 0.4504             \\
\textbf{Amazon}    & -             & 1.0             & 0.6513        & 0.5614             \\
\textbf{IMDB}      & -             & -               & 1.0           & 0.6139             \\
\textbf{TweetEval} & -             & -               & -             & 1.0                \\ \hline
\end{tabular}%
}
\caption{Jaccard similarities between binary vectors $\vq$ trained on different source domains.}
\label{tab:binary_vector_Jaccard_similarity}
\end{table}

\begin{table}[h]
\centering
\resizebox{\columnwidth}{!}{%
\begin{tabular}{lcccc}
\hline
\multicolumn{5}{c}{\textbf{SocialDial}}                                                                                                                                                                                                                                                                                                                                      \\ \hline
\multicolumn{1}{l|}{\textbf{Models}}                      & \textbf{\begin{tabular}[c]{@{}c@{}}Loc (Synthetic)  \\ $\rightarrow$  Loc (Human)\end{tabular}} & \textbf{\begin{tabular}[c]{@{}c@{}}SD (Synthetic)  \\ $\rightarrow$  SD(Human)\end{tabular}} & \textbf{\begin{tabular}[c]{@{}c@{}}SR (Synthetic)  \\ $\rightarrow$  SR(Human)\end{tabular}} & \textbf{Avg- F1} \\ \hline
\multicolumn{1}{l|}{\chatyuan w/o \ourmethod}    & 19.12*                                                                                          & 37.75*                                                                                       & 34.07*                                                                                       & 30.31*           \\
\multicolumn{1}{l|}{\ourmethod-CY}                 & \textbf{23.22}                                                                                  & \textbf{46.04}                                                                               & \textbf{42.71}                                                                               & \textbf{37.32}   \\
\multicolumn{1}{l|}{\ourmethod-CY w/o $\vm$}       & 22.47*                                                                                          & 41.86*                                                                                       & 38.95*                                                                                       & 34.43*           \\
\multicolumn{1}{l|}{\ourmethod-CY w/o $\va$}       & 21.05*                                                                                          & 39.88*                                                                                       & 37.28*                                                                                       & 32.73*           \\
\multicolumn{1}{l|}{\ourmethod-CY w/o $\mathcal{L}_{dist}$}      & 20.17*                                                                                          & 39.26*                                                                                       & 39.41*              
& 32.95*          
\\ 
\multicolumn{1}{l|}{\bart-zh w/o \ourmethod}  & 15.86*                                                                                          & 35.92*                                                                                       & 31.04*                                                                                       & 27.61*           \\
\multicolumn{1}{l|}{\ourmethod-\bart-zh}  & 19.94*                                                                                          & 41.39*                                                                                       & 39.27*                                                                                       & 33.53*           \\
\multicolumn{1}{l|}{\bert-zh w/o \ourmethod}  & 10.34*                                                                                          & 30.17*                                                                                       & 19.87*                                                                                       & 20.12*           \\ 
\multicolumn{1}{l|}{\ourmethod-\bert-zh}  & 14.68*                                                                                          & 36.75*                                                                                       & 27.41*                                                                                       & 26.28*           \\ 
\hline
\end{tabular}%
}
\caption{Ablation study on SocialDial datasets. CY represents the pre-trained language model \chatyuan.}
\label{tab:ablation_study_multiclass_socialdial_cls}
\end{table}


\subsection{Additional Experimental Results}
\label{sec:add_exp}
\begin{table}
\centering
\resizebox{\columnwidth}{!}{%
\begin{tabular}{lccc}
\hline
\multicolumn{4}{c}{\textbf{AG News}}                                                                                                            \\ \hline
\multicolumn{1}{l|}{\textbf{Models}}                  & \textbf{Title $\rightarrow$ Desc} & \textbf{Desc $\rightarrow$ Title} & \textbf{Avg-F1} \\ \hline
\multicolumn{1}{l|}{\bart w/o \ourmethod}    & 80.91*                            & 73.89*                            & 77.40*          \\
\multicolumn{1}{l|}{\ourmethod-\bart}                 & \textbf{89.40}                    & \textbf{81.97}                    & \textbf{85.68}  \\
\multicolumn{1}{l|}{\ourmethod-\bart w/o $\vm$}       & 83.29*                            & 77.08*                            & 80.19*          \\
\multicolumn{1}{l|}{\ourmethod-\bart w/o $\va$}       & 82.72*                            & 77.27*                            & 79.99*          \\
\multicolumn{1}{l|}{\ourmethod-\bart w/o $\mathcal{L}_{dist}$}          & 87.79*                            & 79.82*                            & 83.81*          \\ 
\multicolumn{1}{l|}{T5 w/o \ourmethod}    & 78.48*                            & 71.26*                            & 74.87*          \\
\multicolumn{1}{l|}{\ourmethod-T5}    & 86.91*                            & 79.75*                            & 83.33*          \\
\multicolumn{1}{l|}{\bert w/o \ourmethod}  & 75.12*                            & 61.47*                            & 68.29*          \\
\multicolumn{1}{l|}{\ourmethod-\bert} & 84.79*                            & 75.38*                            & 80.09*          \\
\hline
\end{tabular}%
}
\caption{Ablation study on AG News dataset. 'Binary' refers to the application of the proposed binary classification method on multi-label classification tasks.  }
\label{tab:ablation_study_multiclass_cls}
\end{table}

\end{document}